\documentclass{article}

\PassOptionsToPackage{numbers, compress}{natbib}
\usepackage[final]{neurips_data_2023}

\usepackage[utf8]{inputenc} 
\usepackage[T1]{fontenc}    
\usepackage{hyperref}       
\usepackage{url}            
\usepackage{booktabs}       
\usepackage{amsfonts}       
\usepackage{nicefrac}       
\usepackage{microtype}      
\usepackage{xcolor}         
\usepackage{graphicx}
\usepackage{amsmath}
\usepackage{amssymb}
\usepackage{multirow}
\usepackage{dsfont}
\usepackage{float}
\usepackage{subcaption}

\newcommand{\pt}{\emph{Perception Test}}
\newcommand{\eg}{\emph{e.g. }}

\title{\pt: A Diagnostic Benchmark for Multimodal Video Models}

\author{%
  Viorica P{\u a}tr{\u a}ucean$^1$\thanks{Corresponding author: \texttt{viorica@google.com}, $^1$shared senior contribution} \\
  DeepMind\\
  \And
  Lucas Smaira \\
  DeepMind\\
  \And
  Ankush Gupta \\
  DeepMind\\
  \And
  Adrià Recasens Continente \\
  DeepMind\\
  \And
  Larisa Markeeva \\
  DeepMind\\
    \And
    Dylan Banarse \\
    DeepMind\\
        \And
Skanda Koppula \\
DeepMind\\
\And
Joseph Heyward \\
DeepMind\\
    \And
    Mateusz Malinowski \\
    DeepMind\\
    \And
    Yi Yang \\
    DeepMind\\
    \And
    Carl Doersch \\
    DeepMind\\
    \And
    Tatiana Matejovicova \\
    DeepMind\\
    \And
Yury Sulsky \\
DeepMind\\
    \And
Antoine Miech \\
DeepMind\\
    \And
Alex Frechette \\
DeepMind\\
    \And
    Hanna Klimczak \\
    DeepMind\\
    \And
    Raphael Koster \\
    DeepMind\\
    \And
    Junlin Zhang \\
    DeepMind\\
    \And
Stephanie Winkler \\
DeepMind\\
    \And
    Yusuf Aytar \\
    DeepMind\\
    \And
    Simon Osindero \\
    DeepMind\\
    \And
    Dima Damen \\
    University of Bristol \\
    \And
    Andrew Zisserman \\
    University of Oxford, DeepMind\\
    \And
    João Carreira$^1$ \\
    DeepMind \\
}

\begin{document}

\maketitle

\begin{abstract}
  We propose a novel multimodal video benchmark -- the {\pt} -- to evaluate the perception and reasoning skills of pre-trained multimodal models (e.g.\ Flamingo, SeViLA, or GPT-4). Compared to existing benchmarks that focus on \textit{computational tasks} (e.g. classification, detection or tracking), the \pt\ focuses on \textit{skills} (Memory, Abstraction, Physics, Semantics) and \textit{types of reasoning} (descriptive, explanatory, predictive, counterfactual) across video, audio, and text modalities, to provide a comprehensive and efficient evaluation tool. The benchmark probes pre-trained models for their \emph{transfer} capabilities, in a zero-shot / few-shot or limited finetuning regime. For these purposes, the \pt\ introduces 11.6k real-world videos, 23s average length, designed to show perceptually interesting situations, filmed by around 100 participants worldwide. The videos are densely annotated with six types of labels (multiple-choice and grounded video question-answers, object and point tracks, temporal action and sound segments), enabling both language and non-language evaluations. The fine-tuning and validation splits of the benchmark are publicly available (CC-BY license), in addition to a challenge server with a held-out test split. Human baseline results compared to state-of-the-art video QA models show a substantial gap in performance (91.4$\%$ vs 46.2$\%$), suggesting that there is significant room for improvement in multimodal video understanding. Dataset, baseline code, and challenge server are available at \url{https://github.com/deepmind/perception_test}
\end{abstract}

\section{Introduction}
\label{sec:intro}

Significant progress in multimodal models has been made recently due to large-scale training on multimodal data.  Models like Flamingo~\cite{flamingo}, SeViLA~\cite{yu2023self}, BEiT-3~\cite{beit3}, GPT-4~\cite{openai2023gpt4} show remarkable versatility, dealing with diverse data sources and tackling new tasks by observing only a handful of examples. This is a major departure from specialised models that are typical in computer vision, \eg image or action classifiers~\cite{2022arXiv220104288Y,dosovitskiy2020vit}, object detectors~\cite{Dai_2021_CVPR}, or object trackers~\cite{transtrack}, opening up the path towards general perception and reasoning models.

Benchmarking these models in a robust and efficient way is key to expanding their capabilities, by allowing researchers to rank model design and training choices, and identify areas for improvement.
Many perception-related benchmarks exist, for example Imagenet for image classification~\cite{deng2009imagenet}, Kinetics for video action recognition~\cite{kay2017kinetics}, Audioset for audio event classification~\cite{audioset}, TAO for object tracking~\cite{Dave_2020}, or VQA for image question-answering~\cite{balanced_vqa_v2}, to name only a few. While these benchmarks have led to amazing progress, they all target restricted aspects of perception, focusing on specific computational tasks: \eg image benchmarks discard the temporal dimension, visual question-answering tends to focus on only high-level semantic scene understanding, and object tracking focuses on lower-level, texture-based cues. Gluing several datasets together~\cite{liang2021multibench,schiappa2022robustness} to benchmark more general models (as is done in Flamingo, SeViLA, BEiT-3, or GPT-4) improves coverage, but results in an expensive evaluation procedure that still misses important general perception abilities, \eg physics understanding or memory. Few existing benchmarks even define tasks over both audio and visual modalities~\cite{Grauman2021Ego4DAT}, much less more complex combinations of modalities and tasks. Furthermore, most prior work provides large training sets and thus benchmark models for in-dataset capabilities.

In this work, we propose the {\pt} -- a benchmark formed of purposefully designed, filmed, and annotated real-world videos that aims to comprehensively assess the capabilities of multimodal perception models across different skill areas (Memory, Abstraction, Physics, Semantics), types of reasoning ~\cite{Yi2020CLEVRER} (\textit{descriptive}, \textit{explanatory}, \textit{predictive}, and \textit{counterfactual}), and modalities (video, audio, text). Our benchmark draws inspiration from diagnostic synthetic datasets like CATER~\cite{girdhar2020cater} or CLEVRER~\cite{Yi2020CLEVRER}, behavioral tests like the Visual Turing Test~\cite{malinowski2014towards,VTT}, experiments in developmental psychology~\cite{AguiarBaillargeon02, Baillargeon1994PhysicalRI, parke}, and motor-free perception screening tests used for children or adults~\cite{martin2006test, frostig1965frostig}.

\begin{figure}[h]
    \centering
    \includegraphics[width=0.9\textwidth]{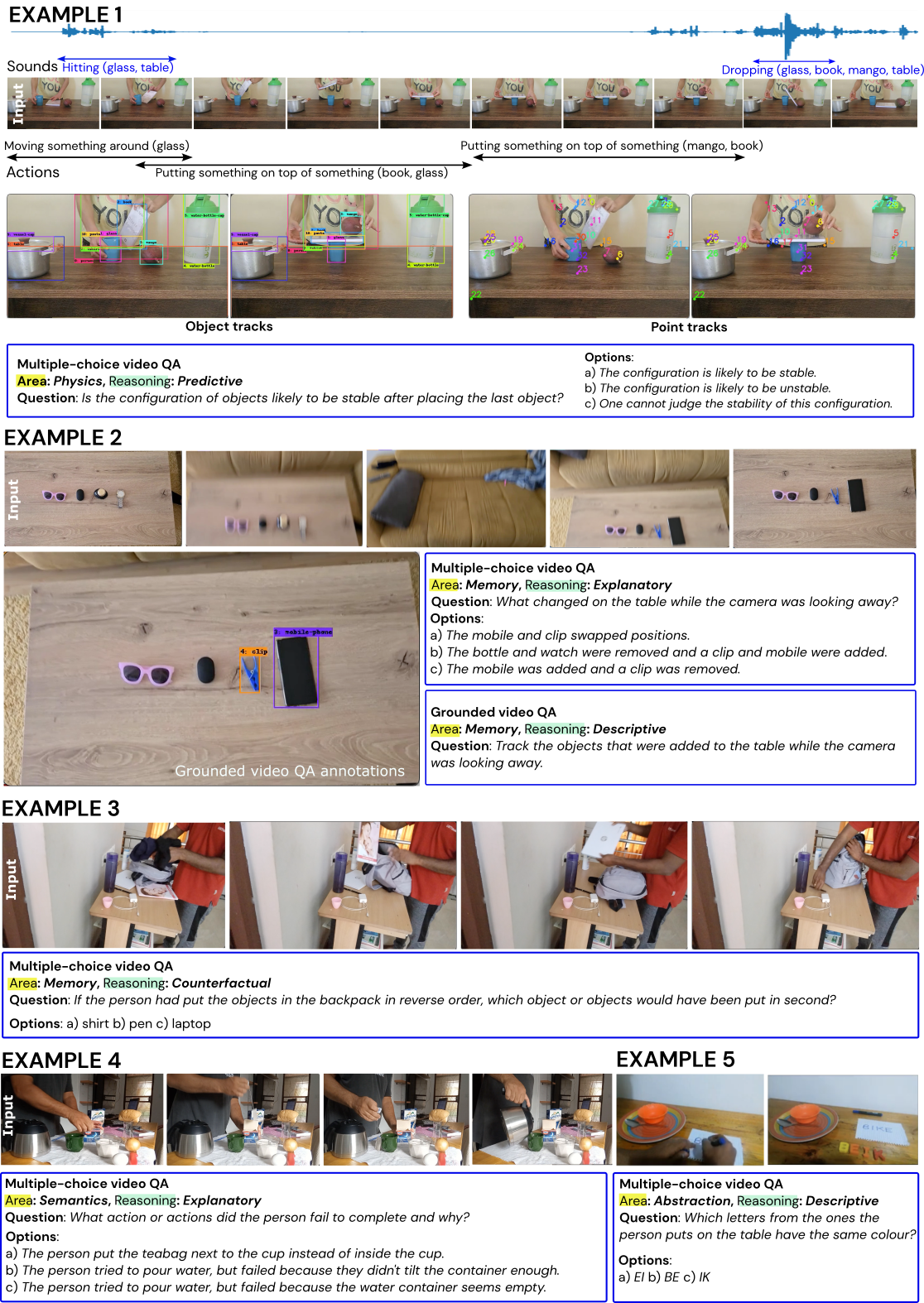} 
    \vspace*{-4pt}
    \caption{\small{The \pt\ contains 6 types of annotations (object \& point tracks, action \& sound segments, multiple-choice videoQA and grounded videoQA) and tasks spanning 4 skill areas (Memory, Asbtraction, Physics, Semantics, and 4 types of reasoning (Descriptive, Explanatory, Predictive, Counterfactual). See the presentation video at \url{https://github.com/deepmind/perception_test} for more examples.}
    }
    \vspace*{-10pt}
    \label{fig:teaser-full}
\end{figure}

To avoid benchmark overfitting, we propose a generalisation-focused evaluation regime. We aim to benchmark any representation or model, pre-trained with any \textit{external} dataset or task, of any scale available.  
The \pt\ itself contains a small training set that can optionally be used for fine-tuning task decoders or prompting the model, and the rest is used for evaluation (public validation and held out test sets). In this regime, we can more robustly assess the \textit{transfer} abilities of these models, such that improvement on the benchmark can more reliably predict generalisation to real-world operation.

The dataset contains 11.6K real-world videos, densely annotated with  
190K object and 8.6K point tracks, 73.5K temporal action segments, 137K temporal sound segments, 38K multiple-choice video question-answer (mc-vQA) pairs and 6K grounded video question-answer (g-vQA) pairs, enabling both language and non-language evaluations, to ensure a thorough assessment; see Figure~\ref{fig:teaser-full} and Table~\ref{tab:annotations}. Having multiple types of annotations per video is useful also for analysis and explainability purposes, as the correlations between successes and failures across tasks may uncover model biases.

We open-source the videos and annotations in the training and validation splits. An evaluation server is made available together with the videos from the held-out test split. Since currently there is no model that can tackle all the evaluation tasks in our benchmark, we provide baseline results for per-task models: object tracking, point tracking, temporal action localisation, temporal sound localisation, multiple-choice video question-answering, and grounded video question-answering. For the mc-vQA task, the performance is mapped across skill areas (memory, abstraction, physics, semantics), and types of reasoning (descriptive, explanatory, predictive, counterfactual) to obtain a comprehensive diagnostics report.  

In the next section (section~\ref{sec:rw}), we discuss related work in more detail, highlighting what sets the \pt\ apart in the rich landscape of multimodal benchmarks. In sections~\ref{sec:videos} and ~\ref{sec:annotations}, we describe the videos and annotations in the \pt, with details about the diversity of participants involved in filming the videos. In section~\ref{sec:ct}, we introduce the computational tasks enabled by these annotations, together with evaluation metrics and baselines, including a human baseline. We conclude with a summary and directions of future work in section~\ref{sec:conclusion}. 

\section{Related work}
\label{sec:rw}

A large number of perception-related benchmarks exist in the literature, covering various computational tasks or modalities. We focus the discussion here on video benchmarks and highlight the differences between the \pt\ and prior work, in terms of data collection process, covered modalities, and available annotations and tasks. 

Existing real-world benchmarks rely on one of the following data sources: \textbf{(i)} Videos collected from the web or repositories like Youtube, \eg Kinetics~\cite{kay2017kinetics}, ActivityNet~\cite{Heilbron_2015_CVPR}, VGGSound~\cite{vggsound}, HVU~\cite{hvu}, ActivityNet-QA~\cite{yu2019activityqa}, tGIFQA~\cite{jang-IJCV-2019}; \textbf{(ii)} Videos collected on demand, filmed by volunteers doing arbitrary activities in indoor or outdoor scenes, \eg EPIC-KITCHENS~\cite{Damen2021TheED}, Ego4D~\cite{Grauman2021Ego4DAT}; \textbf{(iii)}~Videos collected on demand, filmed by crowd-sourced participants doing actions described in pre-defined scripts, mostly in indoor scenes, \eg Charades~\cite{Sigurdsson2016HollywoodIH}, Something-Something v2 (SSv2)~\cite{goyal2018evaluating}. Invariably, all real-world benchmarks use crowd-sourced annotations to enable various computational tasks like action classification, object detection, or video captioning, to name only a few. 

Annotating publicly available videos is useful for training. However, using this approach for general perception evaluation has multiple drawbacks. Large quantities of data would need to be amassed and carefully filtered and annotated to accumulate (statistically) sufficiently diverse samples showing perceptually interesting situations that require skills like memory, abstraction, physics, and semantics understanding. In addition, some types of data are simply not available, \eg situations showing incorrect execution of simple tasks like tying shoe laces. As we aim to assess more diverse skills, we chose to design video scripts that show perceptually interesting and diverse situations and film these with crowd-sourced participants from different places in the world to ensure diversity of video content and appearance. Different from Charades where the scripts were designed by crowd-sourced workers, our scripts are designed by our research team, similar to Something-something (v2). However, we did not aim to obtain an exhaustive coverage of simple actions like in SSv2. Instead, we designed more complex scripts to probe for more advanced reasoning skills beyond action classification.

A few research works have highlighted the need for robust diagnostics benchmarks, \eg  CATER~\cite{girdhar2020cater}, CLEVRER~\cite{Yi2020CLEVRER}, IntPhys~\cite{intphys}, Physion~\cite{physion}. Their authors developed synthetic datasets to evaluate in a more systematic way, across different levels of difficulty, models' abilities to reason about intuitive physics (object collisions, motion, object permanence). We share the same motivation of creating a diagnostic test, and we aim to cover aspects related to memory, abstraction, intuitive physics, and semantics, using real-world videos. To achieve this, in addition to designing the video scripts, our team also designed the questions for each script type for the high-level tasks (mc-vQA and g-vQA); the answers per video were provided by crowd-sourced annotators. Given that our videos are filmed in real world scenes using common household items, the distributions of objects, actions, and sounds in our benchmark have a significant overlap with standard computer vision datasets (\eg 99.01\% of the words in our benchmark also appear in VQAv2~\cite{8100153}), hence the domain gap between the \pt\ and existing large-scale training datasets should be minimal.

Table~\ref{tab:datasets} summarises the characteristics of the \pt\ compared to previous efforts. It can be observed that the \pt\ has a better coverage of skill areas and higher density of annotations\footnote{We count every labeled box, point, temporal segment, or question as a separate annotation}. Size-wise, it is comparable to Charades, but smaller than Ego4D or SSv2. We emphasise that the \pt\ is not designed to be a large-scale training dataset. Instead, it is an evaluation benchmark, with limited fine-tuning or prompting data, meant to assess the transfer capabilities of models.

\begin{table}[t]
    \centering
    \footnotesize
    \begin{tabular}{lcccrrr}
    \hline
    \textbf{Dataset} & \textbf{Source} & \textbf{Skills} & \textbf{\# videos} & \textbf{Dens} & \textbf{L(s)} \\
    \hline
    Charades & C,R & S &  10,000 & 14 & 30\\
    SSv2 & C,R & AS & 108,499 & 1 & 4 \\
    Ego4D-v2 & R &  MS &205,534$^\ddagger$ &9$^*$ &492$^\dagger$ \\
    CLEVRER$^\flat$ & C,Y & P & 60,000 & N/A & 5 \\
    \textbf{\pt} & C,R & MAPS & 11,620  & 761 & 23 \\
    \hline
    \end{tabular}
    \caption{Characteristics of different datasets compared to the \pt. Dataset sources: Scripted (C), Real (R) and Synthetic~(Y). Skill areas: Memory (M), Abstraction (A), Physics (P), Semantics (S). Dens: Average number of annotations per video. L: Average video length in seconds. $^\ddagger$number of annotated clips, $^*$reported for hand-objects subset with the highest density of annotations, $^\dagger$reported for ELM NLQ subset with highest average clip length. $^\flat$: Annotations are extracted directly from the simulator.}
    \vspace*{-8pt}
    \label{tab:datasets} 
\end{table}

\section{Videos in the \pt}
\label{sec:videos}

Inspired by how human perception screening tests are carefully designed by experts in developmental psychology or medicine (\eg~\cite{Cooke2005}), we designed video scripts and tasks to diagnose the perception skills of our models.

\noindent \textbf{Scripts design:}
Our goal was not to obtain an exhaustive coverage of activities or types of scenes. Instead, we selected four areas -- Memory, Abstraction, Physics, Semantics -- within which several skills should be tested (see Table~\ref{tab:skills}, first column) through tasks that require different types of reasoning: descriptive, explanatory, predictive, or counterfactual~\cite{Yi2020CLEVRER}. The skills selection took into account blind spots of existing benchmarks, weaknesses of current models, and aspects that are important for real-world scene understanding.

We then created scripts describing simple situations or games that can be easily performed by any one person (non-professional actor) using the items available in a regular household, or items that can be easily crafted if not available (\eg letters or geometric shapes crafted from paper or cardboard). Each script consists of a brief description of the scene, followed by a description of the actions to be performed, together with specification of the camera placement (static camera one viewpoint; static camera 2 viewpoints; static camera and moving camera). To enhance content diversity, each script had considerable room for variability in the number of objects to be included in the scene or types of actions to be performed, or order of actions.

We prioritised situations where we can test high-level concepts like memory through low-level tasks like object tracking and the other way around: low-level physics understanding probed through high-level tasks like question-answering. In addition, we included in each script elements that could make the situations more interesting and challenging. For example, in cooking scripts (\eg making tea, making salad), we added \textit{distractor actions}~\cite{Sigurdsson2016HollywoodIH}, i.e. actions not relevant for making tea and that have no impact on the outcome of the making tea sequence, like clapping hands, or hitting a kettle with a spoon; this allows probing for understanding of causal relations between actions. We also included \textit{distractor objects} in the scene description, i.e. objects that are not relevant for the current script, but which are relevant for other scripts, like tomatoes present on the table during the make tea activity~\cite{DVN-DMT0PG_2019}. For all the scripts, we also asked participants to include in the scene some  \textit{adversarial configurations of objects} \eg a shoe on the table. This allows us to probe models for understanding of spatial relations of objects when the language biases are not valid. Finally, some of the script variations include \textit{adversarial actions}~\cite{goyal2018evaluating}, i.e. incorrectly executed actions. For example, when making the tea, all the steps are done normally, but one is incorrectly executed, like pouring water from an empty kettle. In this way, we can probe for understanding of task completion, in a more complex setup than the adversarial action classification used in SSv2 dataset~\cite{goyal2018evaluating}.

\begin{table*}[ht!]
    \centering
    \footnotesize
    \begin{tabular}{p{0.23\linewidth}  p{0.75\linewidth}}
    \hline
     \textbf{(Skill Area) Skill} & \textbf{Example of situations and questions or tasks} \\
     \hline
    (\textbf{M})Visual discrimination & Objects are shown in front of the camera, with some shown more than once. \textbf{Task}: Detect which objects were shown multiple times. \\
    \hline
    (\textbf{M}) Change detection &  The camera is filming a table, then looks away for a few seconds, then looks back at the table. Some changes may have occurred. \textbf{Task}: Explain what changed.  \\
    \hline
    (\textbf{M}) Sequencing & Objects are put in a backpack. \textbf{Task}: List their order.  \\
    \hline
    (\textbf{M}) Event recall  & A person indicates a region on the table with the hand, then puts objects inside and outside the region. \textbf{Task}: List the objects put inside the region.  \\
    \hline
    \hline
    (\textbf{A}) Object, action \& event counting  & A person turns a lamp on and off. \textbf{Task}: Count the number of times the illumination changed in the scene. \\
    \hline
    (\textbf{A}) Feature matching  & A person puts wooden letters on the table. \textbf{Task}: Which letters have the same colour?\\
    \hline
    (\textbf{A}) Pattern discovery  & Geometric shapes are shown in a pattern. \textbf{Task}: What shape will be shown next? \\
    \hline
     (\textbf{A}) Pattern breaking  & A person puts multiple cups all facing upwards and one facing downwards. \textbf{Task}: Indicate the object that breaks the pattern. \\
    \hline
    \hline
    (\textbf{P}) Object permanence  & A person plays a cups-game with 3-4 cups by hiding a small object under one of the cups, then shuffles the cups. \textbf{Task}: Predict where is the hidden object after shuffling. \\
    \hline
     (\textbf{P}) Spatial relations  \& containment & A person puts a bookmark in a book, then puts the same or another book in a backpack. \textbf{Task}: Where is the bookmark at the end?\\
    \hline
     (\textbf{P}) Object attributes  & A person writes on a piece of paper. \textbf{Task}: Is the paper lined or plain? \\
    \hline
    (\textbf{P}) Motion \& occluded interactions  & A person moves an occluder object in front of a small object, sometimes moving also the small (occluded) object. \textbf{Task}: Was the small object moved?  \\
    \hline
     (\textbf{P}) Solidity \& collisions & A person launches objects against a blocker object, sometimes removing the blocker. \textbf{Task}: Does the object fall off the table? \\
    \hline
    (\textbf{P}) Conservation  & A person pours an equal amount of water in 2 identical glasses, then pours all or part of the water from one glass in a taller or wider glass. \textbf{Task}: How much water is in the last glass? \\
    \hline
    (\textbf{P}) Stability  & A person puts objects on top of each other in a stable or unstable configuration. \textbf{Task}: Predict if the configuration will be stable after placing the last object. \\
    \hline
    \hline
     (\textbf{S}) Distractor actions \& objects & A person makes tea, and does also some distractor actions unrelated to making tea, \eg rotating a knife. \textbf{Task}: Identify the distractor action(s).\\
    \hline
    (\textbf{S}) Task completion \& adversarial actions  & A person ties shoe laces, but sometimes pretends to tie, or ties the lace of one shoe to the lace of the other shoe. \textbf{Task}: Detect if the action is done correctly. \\
\hline
     (\textbf{S}) Object \& part recognition  & A person conceals a small object in one of their hands, then shuffles the hands. \textbf{Task}: Identify in which hand is the object held. \\
    \hline
     (\textbf{S}) Action \& sound recognition  & All scripts. \textbf{Task}: Detect the actions and sounds in the video from a pre-defined list. \\
    \hline
     (\textbf{S}) Place recognition  & All scripts. \textbf{Task}: Detect where is the action taking place. \\
    \hline
    (\textbf{S}) State recognition  & A person uses an electric device. \textbf{Task}: Indicate if the device is on.\\
    \hline
    (\textbf{S}) General knowledge \& Language  & Some objects are shown to the camera, some multiple times. \textbf{Task}: Given a list of arbitrary statements or word puzzles, some requiring general knowledge to solve, select the statement that contains a reference to the second distinct object shown. \\
    \hline
    \end{tabular}
    \vspace*{-4pt}
    \caption{Examples of scripts probing for different skills in the four areas in the \pt: \textbf{(M)}:Memory, \textbf{(A)}:Abstraction, \textbf{(P)}:Physics, \textbf{(S)}:Semantics.}
    \vspace*{-12pt}
    \label{tab:skills}
\end{table*}

Table~\ref{tab:skills} and Figure~\ref{fig:teaser-full} show examples of situations included in the scripts to probe for different skills in the different areas and types of reasoning. Note that the videos associated with a script allow defining tasks and questions across multiple skill areas. All-in-all, we designed 37 scripts, each with 2-5 variations, to obtain a diverse dataset. Having multiple variations per script allows us to ask the exact same question with the same set of options, and the correct answer depends on the specific script variation -- in this way, we can avoid language biases in questions that give away the answer~\cite{Kervadec_2021_CVPR}. Examples of videos included in the dataset can be found in the presentation video at \url{https://github.com/deepmind/perception_test}.    

\noindent{\textbf{Video filming:}}
Ensuring diversity of participants and scenes depicted in the videos was a critical consideration when developing the benchmark. Using a crowdsourcing pool, we selected around 100 participants from different countries of different ethnicity and gender and aimed to have a diverse representation within each video script. We include in the appendix details about the self-reported demographics of participants. 
Each script variation was filmed by at least a dozen of different participants, using most often a mobile-phone camera, resulting in high-resolution audio-visual assets. For scripts to be filmed from two different viewpoints, the recording was most often done sequentially by repeating the script; a few participants recorded simultaneously using two filming devices. About $15\%$ of the videos were filmed with a moving camera. Most of the videos were filmed indoors in the living room or kitchen, with a small number being filmed in the bathroom or outdoors (about $1\%$). Most of the activities are performed on a tabletop, but some are also performed on the floor or on a chair.
To avoid privacy concerns, we instructed the participants to not record their faces or voices. This is not a limitation of the dataset since the focus in our scripts is on object interactions. The participants gave their consent for the data to be used, published, and stored for perpetuity. 

\noindent \textbf{Splits:}
The \pt\ contains 11609 videos (with audio), 23s average length. It is divided into a small training split (2184 videos, $\sim20\%$ of the data) that can be used for fine-tuning or prompting, a validation split (5900 videos, $\sim50\%$ of the data), and a held-out test split (3525 videos, $\sim30\%$ of the data) available through the evaluation server. We optimised to obtain a good balance across all annotation types and camera motions across the 3 splits; see section~\ref{a:splits} in the appendix.

\section{Annotations in the \pt}
\label{sec:annotations}
We annotate these videos with six types of annotations to cover low-level and high-level aspects, both spatial and temporal, and enable language and non-language evaluations: object and point tracks, temporal action and sound segments, multiple-choice and grounded video question-answers. We include a summary of the number of annotations in Table~\ref{tab:annotations} and visualisations in Figure~\ref{fig:teaser-full}.

\begin{table}[]
    \centering
    \footnotesize
    \begin{tabular}{l r r r r}
    \hline
    \textbf{Annotation type} & \textbf{\# classes} & \textbf{\# annot} & \textbf{\# videos} &  \textbf{Rate (fps)} \\
     \hline
    Objects tracks  & 5101 & 189940 & 11609 & 1 \\
    Point tracks & NA & 8647 & 145 & 30 \\
    Action segments & 63 & 73503 & 11353 & 30 \\
    Sound segments & 16 & 137128 & 11433 & 30 \\
    mc-vQA & 132 & 38060 & 10361 & NA \\
    g-vQA & 34 & 6086 & 3063 & 1 \\
    \hline
    \end{tabular}
    \quad \\
    \vspace{1mm}
    \begin{tabular}{l r r}
    \hline
     \textbf{Area} & \# videoQA  \\
    \hline
    Memory & 7256 (36) \\
    Abstraction & 12737 (58) \\
    Physics & 23741 (80) \\
    Semantics & 24965 (82) \\
    \hline
    \end{tabular}
    \quad
    \begin{tabular}{l r r}
    \hline
    \textbf{Reasoning}  & \# videoQA \\
    \hline
    Descriptive & 31536 (106) \\
    Explanatory & 4513 \hspace{1mm} (14)\\
    Predictive & 1278  \hspace{2mm} (7) \\
    Counterfactual & 733 \hspace{2mm} (5) \\
    \hline
    \end{tabular}
    \caption{\textbf{Top}: Annotations in the \pt. Each object or point track contains frame-level annotations at a certain \textit{frame rate}, \eg each point is annotated on every frame, at 30 fps. Action and sound segments are annotated at the original video frame rate. \# classes refers to the number of unique object names for object tracks and the number of unique questions for multiple-choice videoQA (mc-vQA) and grounded videoQA (g-vQA). \textbf{Bottom}: Number of videoQA pairs and (unique questions) per area and type of reasoning. Note that one question may be counted in multiple areas if it tests more than one skill. Each question is assigned a unique type of reasoning.}
    \vspace*{-12pt}
    \label{tab:annotations}
\end{table}

\noindent{\textbf{Object tracks:}}
Object tracks represent the \textit{root annotation} of our benchmark. All the other annotations, except for multiple-choice vQA, are linked or grounded into object tracks. In the annotation process, we instructed annotators to focus on the objects that the person interacts with and the objects that are in the immediate vicinity of the area where the person is performing actions, which act as distractor objects. We annotated boxes at 1fps throughout the video, which gives a good trade-off between density of annotations and annotation cost. When the objects are occluded, the annotators marked an approximate position of the boxes. Some ambiguous classes still remain, like liquids being poured or objects being torn. The object names were defined from an open vocabulary. The annotators typically included object attributes as well (colour, material), resulting in a large number of unique names. A list of the most frequent words (object or attributes) is included in the appendix, Fig.~\ref{fig:objects-actions} (left), together with the distribution of object tracks into various categories, \eg objects involved in actions or sounds correlated with camera motion (Table~\ref{tab:tracks}).

\noindent \textit{Cups-game subset:}
We isolate the videos corresponding to the cups-game scripts, as they can be an interesting subset for probing object trackers' abilities to reason about motion, object permanence, or occluded interactions when different factors may influence the difficulty of the task, \eg identical vs non-identical objects used in the game, transparent vs non-transparent objects, or number of objects used. This subset contains 598 videos, with 483 videos where the cups are identical, and 113 videos where the cups are transparent. Most of the videos have 3 cups (451 videos), 132 videos have 2 cups, and 34 videos have 4 cups. We also provide a visibility mask for each video showing when the hidden object is occluded.

\noindent{\textbf{Point tracks:}} Although object tracks based on bounding boxes allow probing some physical properties of objects, such as object permanence, solidity, and coarse motion, they do not fully describe articulated or non-rigid objects, thin objects that are not axis-aligned, or out-of-plane rotation. A better understanding of physical interactions arises if we can track how object \textit{surfaces} move and deform over time.  To this end, we annotate point tracks on object surfaces following the protocol of TAP-Vid~\cite{doersch2022tapvid}. Annotators were instructed to select points spanning all the different parts of the objects labelled in the object tracking task. Points that are occluded are simply marked as occluded and not tracked. For translucent objects (\eg glass cups), we only consider points to be `visible' if they belong to the surface closest to the camera. The annotated points are dense in time (30fps). Table~\ref{tab:points} in the appendix gives the distribution of points that are moving or static, as well as those on videos with moving cameras.

\noindent{\textbf{Action segments with action-relevant objects:}}
To capture temporal understanding and enable grounding over time, we annotate the videos with temporal segments belonging to a fixed set of templated labels, \eg \textit{putting something into something}, similar to~\cite{goyal2018evaluating}. These are associated with action-relevant object tracks, i.e.\ objects involved in the action.   The action boundaries are defined based on contact with action-relevant  objects. For instance, when a person puts sugar in a tea, the \textit{putting something into something} action starts when the person picks up the spoon and ends when the person puts down the spoon. If, after putting the sugar, the person starts stirring with the same spoon, this defines a new segment as the type of action changed. The frequency of actions across the entire dataset is included in the appendix, Fig.~\ref{fig:objects-actions} (right).

\noindent{\textbf{Sound segments with sound-relevant objects:}}
Similarly to the action segment annotations but applied to the audio modality, we collect sound segment annotations grounded in object tracks. By watching the video and listening to the audio, the annotators define temporal sound segments and label them from a list of 16 audio segment labels. For each sound, the annotators also identify the object (or objects) involved in making the sound, or specify that these are out of the camera's field of view.
For example, if an object is placed on the table making an audible sound, then both the object track and the table track are associated with the sound segment.
The frequency of sounds across the entire dataset is included in the appendix, Fig.~\ref{fig:sounds}.

\noindent{\textbf{Question-answers for video-level reasoning:}}
Different from the existing VQA datasets, which rely on crowd-sourced questions and answers, our team designed the questions per script to cover different types of reasoning~\cite{Yi2020CLEVRER}: descriptive, explanatory, predictive, counterfactual, and to cover aspects that are important for operating in the real world, \eg understanding task completion, detecting changes, and so on. The answers for all the questions per video were provided by crowd-sourced participants. As we are interested in non-ambiguous evaluation, we favour the multiple-choice setup over the open-language answer setup. To define challenging negative options, we partly relied on human annotators, partly sampling from the correct answers of other videos in the same type of script. 
Table~\ref{tab:annotations} bottom and Figure~\ref{fig:mcqa} in the appendix show the distribution of question-video pairs into perception skills, skill areas, and type of reasoning.

\noindent{\textbf{Question-answers with answer-relevant objects:}}
As another way to connect high-level and low-level scene understanding capabilities, we define questions or tasks in language form, with answers given as object tracks. Similar to the multiple-choice question-answers above, our team defined the questions, and human raters selected the answers from the existing object tracks. The grounded questions are associated with skill areas and types of reasoning.

\section{Computational tasks and baseline results in the \pt}
\label{sec:ct}
\noindent{\textbf{Computational tasks:}}
We defined six computational tasks based on the annotations available in the \pt. We summarise in Table~\ref{tab:tasks} the task definitions (outputs, metrics) and the performance of top-performing baselines. It can be observed that the \pt\ combines lower-level dense prediction tasks like object and point tracking, whose outputs are box and point trajectories, with higher-level tasks like video question answering. For all the tasks, the video and audio are available as inputs, together with a task specification where applicable, \eg the coordinates of a box to track for object tracking, or a language question and options for multiple-choice videoQA. More details about the task definitions are included in the appendix. Note that many other computational tasks can be defined based on the available annotations, \eg grounded temporal action/sound localisation.

\begin{table}[t]
    \centering
    \footnotesize
       \begin{tabular}{lcccc}
    \hline
     \textbf{Task}& \textbf{Output} & \textbf{Metric} & \textbf{Baseline} & \textbf{Score}\\
     \hline
    Object tracking & box track &  Avg. IoU & SiamFC~\cite{bertinetto2016fully} & 0.67 \\
    Point tracking & point track &  Avg. Jaccard & TAP-Net~\cite{doersch2022tapvid} & 0.401 \\
    Temporal action localisation &  list of action segments & mAP & ActionFormer~\cite{zhang2022actionformer} & 15.56 \\
    Temporal sound localisation & list of sound segments& mAP & ActionFormer~\cite{zhang2022actionformer} & 15.46 \\
    multiple-choice videoQA & answer (1 out of 3) &  top-1 accuracy & SeViLA~\cite{yu2023self} & 46.2 \\
    grounded videoQA & list of box tracks &  HOTA~\cite{luiten2020IJCV}  & MDETR~\cite{kamath2021mdetr}+Stark~\cite{Yan2021LearningST} & 0.1 \\
    \hline
    \end{tabular}
    \caption{Computational tasks and top-performing baselines in the \pt: the model receives a video with audio, plus a task-specific input (\eg the coordinates of a bounding box for the object tracking task), and produces a task-specific prediction, evaluated using dedicated metrics.}
    \label{tab:tasks}
    \vspace*{-12pt}
\end{table}

\noindent{\textbf{Baselines:}} Ideally, a single model should be able to perform all the tasks in the \pt. Since such a model is not available in the literature, we include results obtained with per-task baselines on the validation split for all the six tasks in the \pt; see Table~\ref{tab:tasks} for a summary of top-performing baselines and their average performance, and the appendix for more details. When selecting these baselines, we favoured strong-performing models that can be evaluated in a zero-shot or few-shot setting, as our focus is on generalisation. However, for action and sound localisation, such models do not exist in the literature, so fine-tuning on our set of classes was necessary. For the mc-vQA task, we also provide a human baseline and fine-tuned evaluation to further assess the difficulty of the dataset for humans and for SOTA video-language models, respectively.

\noindent \textbf{Object tracking:} The overall performance of SiamFC~\cite{bertinetto2016fully} (UniTrack~\cite{wang2021different} implementation) on our benchmark confirms the findings from~\cite{fan2021lasot} that simple Siamese trackers are better when probed zero-shot than more complex recent trackers, \eg Stark tracker~\cite{Yan2021LearningST} obtains 0.56 mean IoU on the \pt\ vs 0.67 for SiamFC. Even for SiamFC, the tracking performance drops when the camera and/or the objects are moving. The results for the different categories of objects (involved in actions or in sounds, etc.) are included in Table~\ref{tab:box-results}, aggregated based on camera motion.

\noindent \textbf{Point tracking:} The performance of our baseline TAP-Net~\cite{doersch2022tapvid} is a bit lower on the \pt\ compared to the performance reported by the authors on the Kinetics dataset~\cite{kay2017kinetics} (0.466 vs 0.401); see detailed results in Table~\ref{tab:point-results-details}. We attribute this drop in performance mainly to the increased video length in our benchmark (23s in \pt\ compared to 10s in Kinetics).

\noindent \textbf{Action localisation:} The confusion matrix for our fine-tuned baseline ActionFormer~\cite{zhang2022actionformer} (Fig.~\ref{fig:cm}) shows that the model struggles mostly with rare action classes that are confused with more frequent ones, and it also confuses pretend actions with their non-pretend versions, \eg \textit{ironing something} vs \textit{pretending to iron something}. Using multimodal inputs does not increase the performance significantly, as summarised in Table~\ref{tab:action-sounds-map}, top. Overall, ActionFormer's performance on our benchmark is lower  compared to other benchmarks (15.56 mAP on \pt\ vs 22.7 mAP on EPIC-Kitchens~\cite{Damen2018EPICKITCHENS}), most likely due to the presence of adversarial actions and our limited training set. We hope to see in the near future models that can handle open-vocabulary action classes (similarly to open-vocabulary object detection~\cite{Kaul2023}), so that fine-tuning is no longer necessary. 

\noindent \textbf{Sound localisation:} We adapted the same ActionFormer model~\cite{zhang2022actionformer} to perform the localisation task in the audio modality. The best performance is obtained when features from both video and audio modalities are used as input; see Table~\ref{tab:action-sounds-map}, bottom.

\noindent \textbf{Multiple-choice vQA:} We report results for two strong recent video language models: Flamingo~\cite{alayrac2022flamingo} in zero-shot and few-shot setups, and SeViLA~\cite{yu2023self} in zero-shot and fine-tuned regimes. We also include a dummy frequency-based baseline and a human baseline. For the frequency baseline, given that each question-options pair is defined over multiple videos, we keep as answer the option that is most frequently the correct answer in the training set. One can also compute this baseline on a random subset of training examples for each question, see Table~\ref{tab:vqa_comparison}, to obtain a fairer dummy baseline for models using few-shot evaluation.

\begin{figure}[h]
    \centering
    \includegraphics[width=0.9\linewidth]{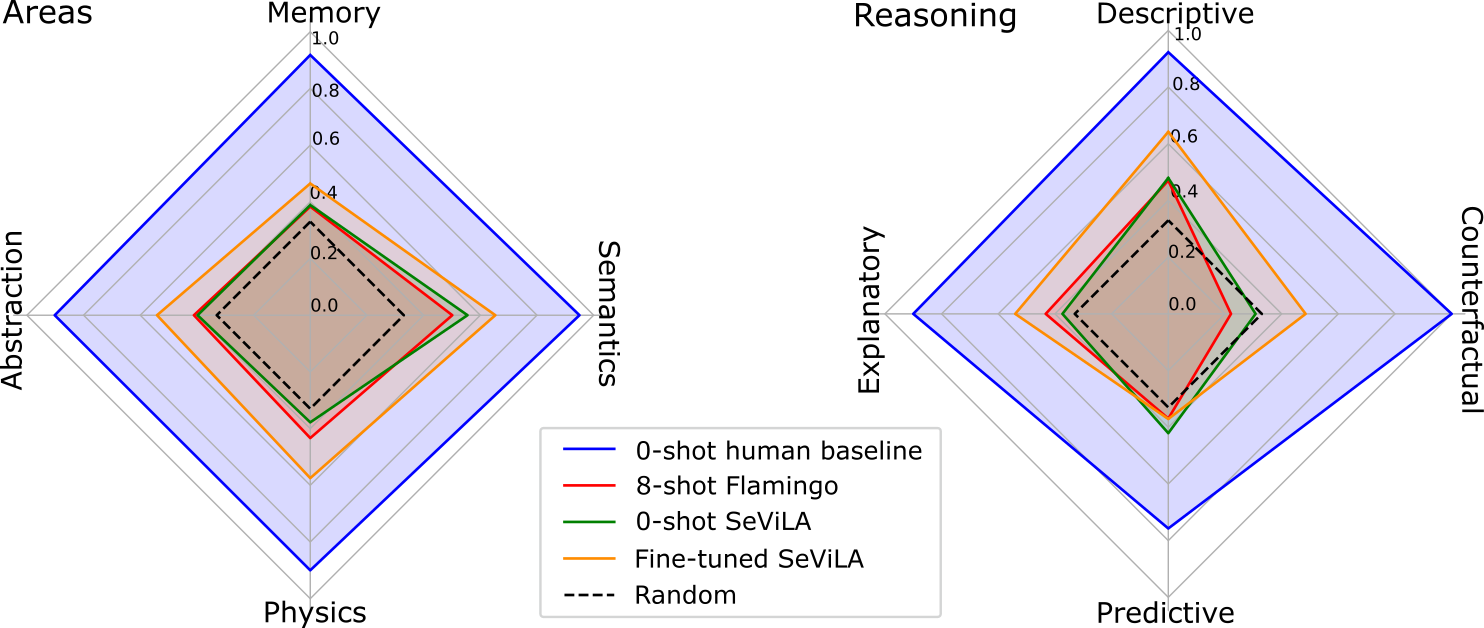}
    \vspace*{-4pt}
    \caption{0-shot human baseline compared to 8-shot Flamingo-3B, 0-shot and fine-tuned SeViLA, and random baseline on the validation set. In 0-shot and 8-shot regimes, both Flamingo and SeViLA are far from the 0-shot human baseline. SeViLA fine-tuning improves results to some extent, but the gap to human performance is still substantial.}
    \label{fig:flamingo-human}
    \vspace*{-12pt}
\end{figure}

\noindent \textit{Human baseline}. We ran a small study for the mc-vQA task with human participants. We used 126 questions from the dataset, with one video per question selected at random. We recruited 30 crowd-sourced participants (half male, half female, with advanced English skills), different from the raters annotating the videos. Each participant answered a subset of 42 questions, resulting in 10 answers per question. The performance per area and type of reasoning is detailed in Figure~\ref{fig:flamingo-human}. The overall average accuracy was $91.4\%$. The mistakes occurred in situations difficult to judge from the given viewpoint, \eg if a configuration of objects would be stable (without seeing the end of the video), or in edge cases where humans overlooked details happening very early on in the video. It is worth noting that the participants did not require any training, which is similar to a zero-shot setup. The median time spent to answer 42 questions was 30 minutes.

\begin{figure}[h]
    \centering
    \includegraphics[width=.97\linewidth]{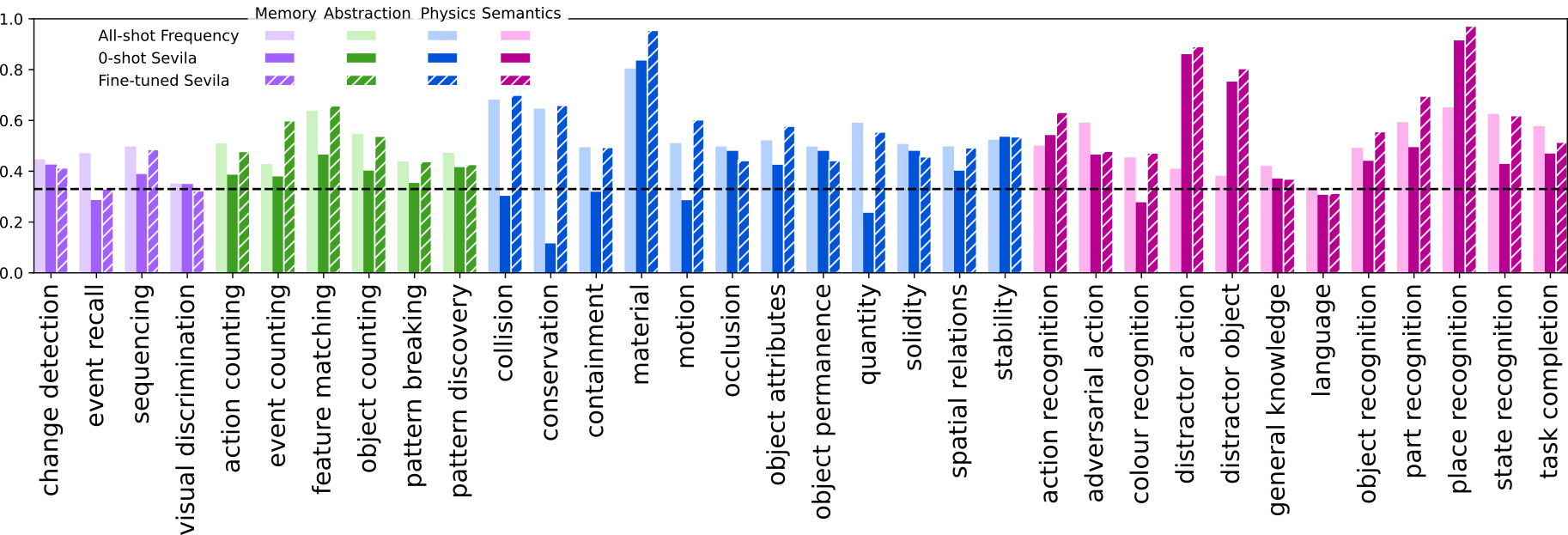}
    \caption{Performance on the validation set across skills for the 0-shot and fine-tuned SeViLA compared to frequency dummy baseline. The black dashed line indicates the random baseline.}
    \label{fig:sevila-skills}
    \vspace*{-8pt}
\end{figure}

It can be observed that both Flamingo and SeViLA are far from human performance when evaluated 0-shot or few-shot and cannot outperform the 8-shot dummy frequency baseline; see Figures~\ref{fig:flamingo-human},~\ref{fig:sevila-skills},~\ref{fig:flamingo-skills}, and Table~\ref{tab:vqa_comparison}. On many skills in the Memory, Physics, and Abstraction areas, their performance is below the 8-shot frequency dummy baseline, and in a few cases, \eg (Piaget) conservation task, collision, or counterfactual reasoning, they are even below the pure random baseline. For counterfactuals, our qualitative investigation shows that Flamingo tends to latch on the visible elements in the video, failing to imagine the alternate reality that counterfactual questions require; \eg for videos where a person writes some letters on the paper, the (counterfactual) question posed is: \textit{What would be the order of the written letters if the person had written them in reverse order?}. Flamingo often selects the written order as correct. Interestingly, the larger versions of the model (due to larger language branches) seem to fare worse overall, which might point to overfitting issues. However, we leave an in-depth analysis for future work. Fine-tuning SeViLA leads to better results compared to all-shot frequency baseline, but mainly in the Semantics area (Fig.~\ref{fig:sevila-skills}). SeViLA's 0-shot/fine-tuned scores on \pt\ (Fig.~\ref{fig:sevila-skills}) are significantly lower than on NExT-QA benchmark~\cite{xiao2021next}: 46.2 vs 63.6 for zero-shot, and 62.0 vs 73.8 on fine-tuned. We attribute this to the diversity of skills and the hard negative options included in the \pt.

\noindent \textbf{Grounded-vQA:} As no existing model can perform this task, we created a baseline by running MDETR~\cite{kamath2021mdetr} on the middle frame and then tracking the predictions using the Stark~\cite{Yan2021LearningST} object tracker. The performance is poor; see Table~\ref{tab:gvqa} and Figure~\ref{fig:hota}. The failures are caused mainly by poor detection results -- since the tasks are temporal in nature, extracting \textit{seed} boxes from the middle frame is not sufficient to solve the tasks, calling for models capable of dealing with both spatial and temporal dimensions.

\section{Conclusion}
\label{sec:conclusion}

We propose a diagnostic benchmark for multimodal models, to probe for memory, abstraction, physics, and semantic capabilities, across visual, audio, and text modalities, using real-world videos purposefully designed and filmed to show interesting perceptual situations. Solving the tasks requires different types of reasoning: descriptive, explanatory, predictive, and counterfactual. The videos are densely labeled with six types of annotations (objects and point tracks, action and sound segments, multiple-choice and grounded video question-answers). We are open-sourcing the videos and the annotations in the train and validation splits, together with per-task baseline results and evaluation code. A challenge server is available to evaluate models on the held-out test split. In principle, any model can be evaluated on our benchmark, either in a zero/few-shot setting or by fine-tuning on our limited train split. An ideal perception model would be able to perform all the tasks in our benchmark. Our results suggest that state-of-the-art zero-shot or few-shot video-language models are not able to outperform a dummy frequency-based baseline, whereas humans in the same setting are nearly perfect. We discuss limitations, and ethical and societal aspects in the appendix. We hope that our work will contribute to understanding models' limitations (through direct evaluation and interpretability analysis supported by the different types of annotations) and narrowing down areas of improvement to guide research towards general perception models.

\section*{Acknowledgments} We are grateful to Luis Piloto, Kenneth Marino, Luyu Wang, Felix Hill, Martin Chadwick, Lucy Campbell-Gillingham, Boxi Wu, Drew Jaegle, Pauline Luc, Marianne Monteiro, Anna Bulanova, Radu Isac, Muqthar Mohammad, Vijay Vibha Tumala, Mahesh Maddinala, Yiwen Luo, Alina Kuznetsova, Aida Nematzadeh, Lisa Anne Hendricks, Aishwarya Agrawal, Nando de Freitas, Matt Botvinick, Shane Legg, and Relja Arandjelovic for providing insightful input on this project.

\medskip

{\small
\bibliographystyle{plainnat}
\bibliography{main}
}

\newpage
\appendix

\section*{\textbf{Appendix}}
\setcounter{table}{0}
\renewcommand{\thetable}{A\arabic{table}}
\pagenumbering{arabic}

\setcounter{figure}{0}
\renewcommand{\thefigure}{A\arabic{figure}}

\renewcommand{\thesection}{\arabic{section}}
\setcounter{section}{0}

\section{\pt\ at a glance}
Figure~\ref{fig:teaser-full} and the presentation video available at \url{https://github.com/deepmind/perception_test} summarise the types of videos, annotations, and tasks available in the \pt.

\section{More details about annotations in the \pt}
\label{sec:annots}

\begin{figure}[h!]
    \centering
    \includegraphics[width=.31\textwidth]{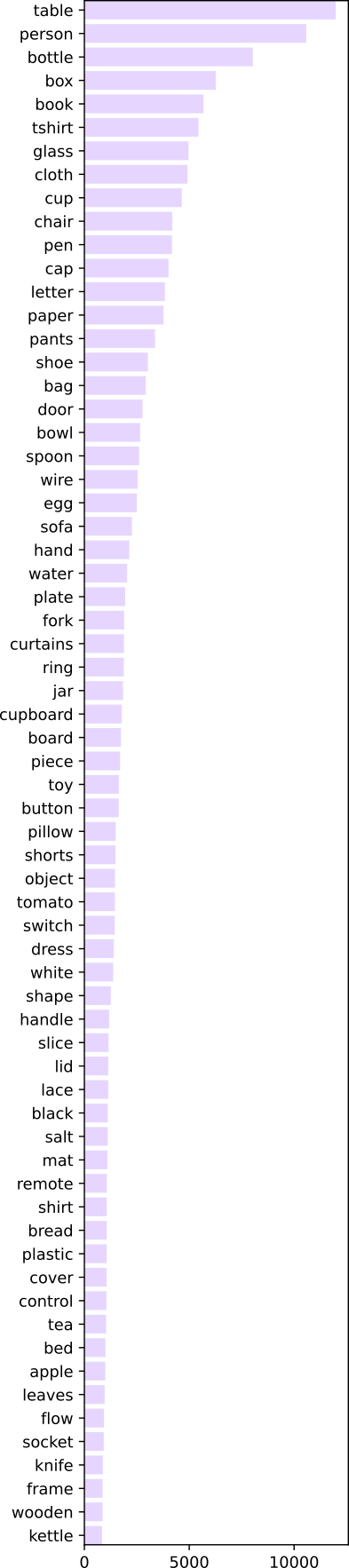} 
    \includegraphics[width=.63\textwidth]{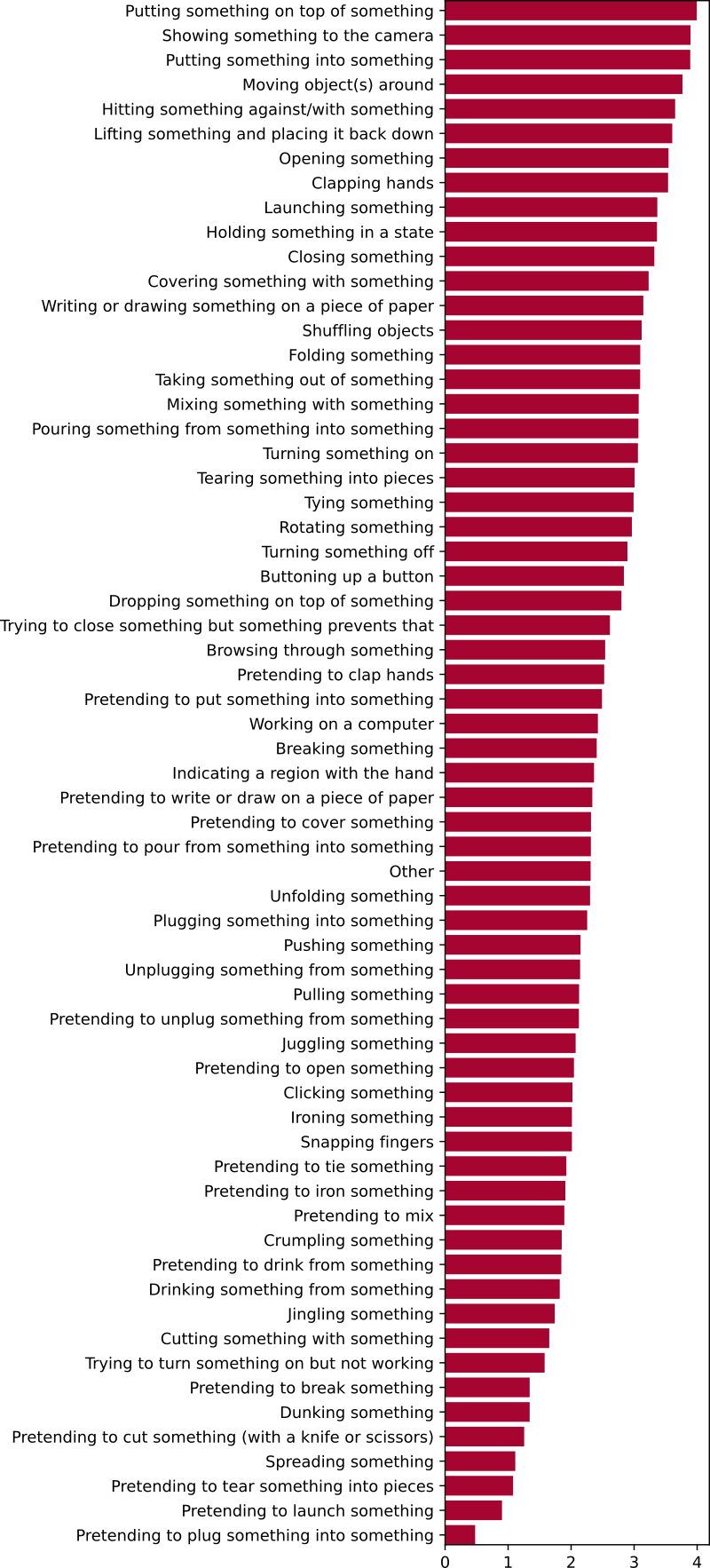} 
    \caption{Frequency of objects and log-scale frequency of actions in the \pt.}
    \label{fig:objects-actions}
\end{figure}

\begin{figure}[h!]
    \centering
    \includegraphics[width=\textwidth]{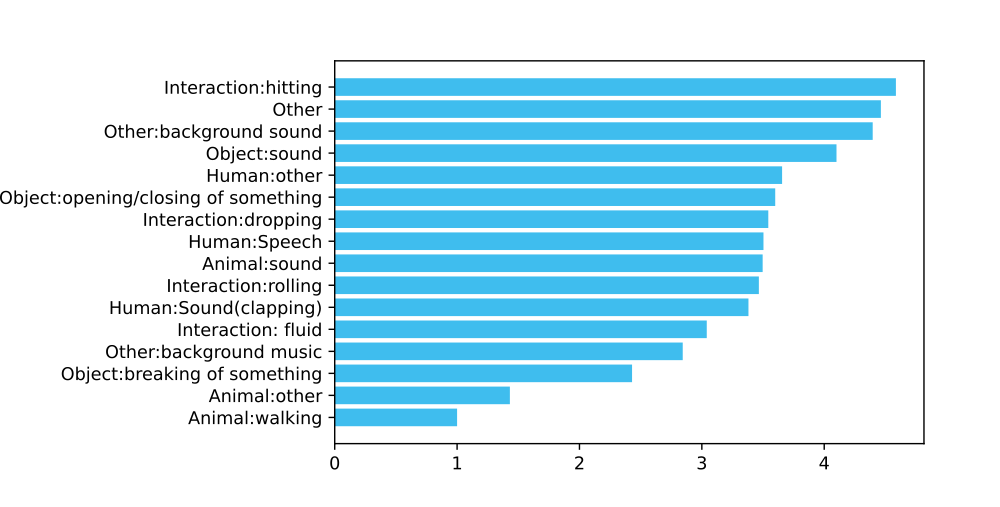} 
    \caption{Log-scale frequency of sounds in the \pt.}
    \label{fig:sounds}
\end{figure}

\begin{figure}[h!]
    \centering
     \includegraphics[width=\textwidth]{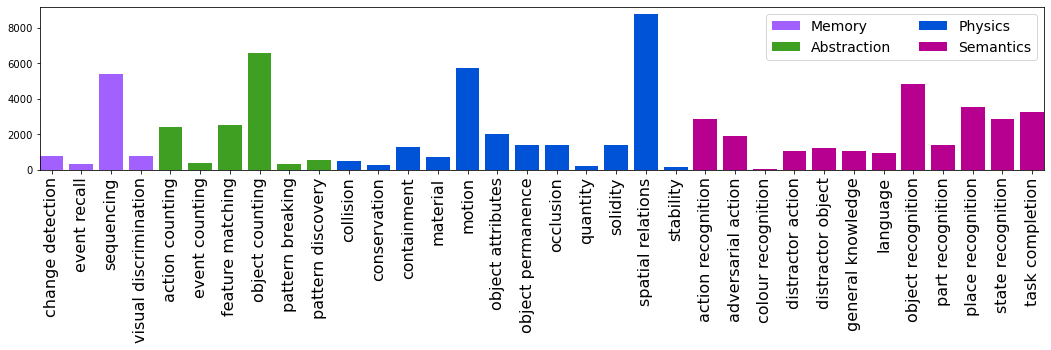} 
    \caption{Number of multiple-choice video question-answers in the \pt\ across skills in the four skill areas: Memory, Abstraction, Physics, Semantics. One skill can be assigned to multiple skill areas---here we choose one as the prime area for each skill.}
    \label{fig:mcqa}
\end{figure}

The distributions of object and point tracks across camera motion and objects involved in actions, sounds, and grounded vQA are included in Table~\ref{tab:tracks} and Table~\ref{tab:points}. Figures~\ref{fig:objects-actions} and~\ref{fig:sounds} present the frequency of popular words included in object names, and the distribution of actions and sounds respectively. Figure~\ref{fig:mcqa} shows the distribution of questions across skills.

\begin{table}[H]
    \centering
    \footnotesize
    \begin{tabular}{l c c r}
    \hline
     \textbf{Camera} & \textbf{Static}  & \textbf{Moving} & \textbf{Total}  \\
    \hline
    \textbf{\# total objects} & 165552 & 26164 &191716 \\  
    \textbf{\# action objects} & 55344 & 6923 &62267 \\
    \textbf{\# sound objects} & 56158 & 7666 &63824 \\
    \textbf{\# g-vQA boxes} & 6795 & 2579 &9374 \\ 
    \hline
    \end{tabular}
    \caption{Object tracks involved in actions, sounds, and grounded-vQA, split by camera motion.}
    \label{tab:tracks}
\end{table}

\begin{table}[h!]
    \centering
    \footnotesize
    \begin{tabular}{l c c r}
    \hline
     \textbf{Camera} & \textbf{Static}  & \textbf{Moving}  &\textbf{Total} \\
    \hline
    \textbf{\# total points} & 7791 & 783 &8574 \\
    \textbf{\# moving points} & 3800 & 783 &4583 \\
    \textbf{\# static points} & 3991 & 0 &3991 \\
    \hline
    \end{tabular}
    \caption{Point tracks available in the \pt, split by point and camera motion.}
    \label{tab:points}
\end{table}

\section{Computational tasks}

\noindent{\textbf{Single object tracking:}} In this task, the model separately tracks every single object labelled in the dataset starting from an initial bounding box. In some cases ($\approx 20\%$) where the object is entering the field of view at the beginning or during the video, the first box may span only a few pixels, so it does not contain a representative view of the object. To deal with this problem, we use a heuristic to select a later frame, when the object is not touching the image boundary, to identify the query box for each object track. Performance is evaluated using the standard \emph{average intersection-over-union} (IoU) metric, (also called average overlap), for evaluating long-term tracking without tracker re-initialization. It is defined as the average IoU over the entire track between the predicted and the ground-truth boxes~\cite{got10k,vcehovin2016visual}.
We also provide code for more fine-grained analysis, \eg performance on objects in videos shot with static vs. moving cameras, objects involved in actions etc.

\noindent \textit{Cups-game subset:} For the occluded object involved in cups-games, we use intersection as a metric for tracking (as opposed to Intersection-over-Union), to deal with the uncertainty of the position when the object is occluded.

\noindent{\textbf{Single point tracking:}} In this task, given a set of ground truth initial 2D point coordinates, the model should separately trace their spatial trajectories throughout the video. Performance is evaluated using the recently proposed \emph{average Jaccard} metric for evaluating both long-term point tracking position and occlusion accuracy. This metric checks how similar the predicted and the ground-truth point tracks are, based on the average number of true positive matches, divided by the sum of true positives, false positives, and false negatives over the entire track~\cite{greff2022kubric,doersch2022tapvid}. 

\noindent{\textbf{Temporal action / sound localisation:}} We define these two tasks similarly, as temporal segment detection problems. Given a video, the model predicts potentially overlapping temporal 1d-segment covering the actions/sounds and classifies them using a fixed set of labels. Performance is evaluated using the standard mean AP over classes~\cite{zhang2022actionformer} based on temporal IoU between predicted and ground truth temporal segments. 

\noindent{\textbf{Multiple-choice video question-answering:}} 
In this task, the model receives, in parallel with the video, a question and three possible answers, out of which only one is correct, and the model has to pick one answer (33\% random chance). 
For most of the questions, watching the video and reading the question are enough for providing a correct answer. A limited number of questions are formulated in a generic way, so the options are necessary for choosing the answer: \eg \textit{Which of the following statements describes the scene better?}. In some cases, choosing the answer by elimination of the false options may be simpler. 
Performance is evaluated by measuring top-1 accuracy. For a couple of scripts, the videos must be trimmed to not reveal the answer: in the cups-games and stable configurations videos, we provide a frame id where the video should be trimmed. For the train and validation splits we release the entire videos together with the cut frame id information. In the held-out test split, only the trimmed videos are available for these particular video types.

\noindent{\textbf{Grounded video question-answering:}} 
This task is similar to conditional multiple-object tracking, with the conditioning given as a language task or question as opposed to a class label~\cite{8100260}. The answers are object tracks defined throughout the video and we use HOTA~\cite{luiten2020IJCV} metrics to evaluate performance. In some situations, the initial parts of the track might not be relevant for the question, \eg \textit{Track the object that was removed from the table} and the object is removed halfway through the video. However, given that we do not enforce causal processing of the video, the track prediction for the initial part can still be done in hindsight. 

\section{Baseline results}
\label{sec:appendix:baselines}

\noindent{\textbf{Object tracking:}}
We report baseline results using the SiamFC model~\cite{bertinetto2016fully} (UniTrack~\cite{wang2021different} implementation). SiamFC was chosen due to its high-performance on a number of single-object tracking benchmarks when running in zero-shot setting~\cite{fan2021lasot}. We also include a static dummy baseline that assumes all objects are static, so it just replicates throughout the video the box-to-track received as input. The results for the different categories of objects (involved in actions or in sounds, etc) are included in Table~\ref{tab:box-results}, aggregated based on camera motion.

\begin{table}[h]
    \centering
    \footnotesize
    \begin{tabular}{l c c c}
    \hline
    \textbf{Object Tracking} & \textbf{All} & \textbf{Static camera}  & \textbf{Moving camera}  \\
    \hline
    \textbf{all objects} & 0.66 / 0.67 & 0.70 / 0.69 & 0.42 / 0.54 \\
    \textbf{action objects} & 0.48 / 0.53 & 0.50 / 0.54 & 0.31 / 0.47 \\
    \textbf{sound objects} & 0.56 / 0.60 & 0.58 / 0.61 & 0.40 / 0.53 \\
    \textbf{g-vQA boxes} & 0.38 / 0.50 & 0.43 / 0.51 & 0.26 / 0.47 \\
    \hline
    \end{tabular}
    \caption{Static dummy baseline / SiamFC results, measured as average IoU, across different categories of objects in the \pt. Since many objects are static, the performance of the dummy baseline is good overall, but it degrades considerably when motion is involved, whereas the SiamFC tracker is more robust.}
    \label{tab:box-results}
\end{table}

\noindent{\textbf{Point tracking:}} 
We report baseline results using a TAP-Net model~\cite{doersch2022tapvid} trained on Kubric~\cite{greff2022kubric} and transferred zero-shot. The model operates on 256x256 resolution (aspect ratio is not preserved) and consumes the whole video directly. We also include a static dummy baseline assuming all future points are visible and never change the location. Table~\ref{tab:point-results-details} shows the results. As expected, both moving points and points seen through a moving camera are considerably harder to track. 
Following~\cite{doersch2022tapvid}, we use three evaluation metrics. (1) \textit{Position Accuracy ($<\delta^x$)}: for a given threshold $\delta$, we measure the fraction of points that are within the threshold of their ground truth, for frames where points are visible. For all predictions, we resize them to 256x256 resolution and measure $<\delta^{x}$ across 5 thresholds: 1, 2, 4, 8, and 16 pixels. (2) \textit{Occlusion Accuracy (OA)}: a simple classification accuracy for the point occlusion prediction on each frame. 
(3) \textit{Jaccard at $\delta$}: an evaluation metric considering both occlusion and position accuracy.  It is the fraction of `true positives', i.e., points within the threshold of any visible ground truth points, divided by `true positives' plus `false positives' (points that are predicted visible, but the ground truth is either occluded or farther than the threshold) plus `false negatives' (groundtruth visible points that are predicted as occluded or the prediction is farther than the threshold). Our final metric \textit{Average Jaccard (AJ)} averages Jaccard across all 5 thresholds: 1, 2, 4, 8, and 16 pixels.

To further understand the performance, we split points into two groups: static and moving. Note that there are no static points in the moving camera scenario, all points are moving. In static camera, we determine that a point is moving if its distance between start frame and end frame is more than 0.01 in the normalized image coordinate system. As expected, the dummy baseline performs well on static points, reaching 0.722 average jaccard. But TAP-Net significantly outperforms when points are moving, particularly in the moving camera setup, improving average Jaccard from 0.088 to 0.328. Besides AJ, TAP-Net significantly improves the static baseline on occlusion accuracy from 0.675 to 0.849. One interesting observation is that on both position accuracy ($<\delta^x$) and jaccard at $\delta$, TAP-Net starts to outperform static baseline only when measured above 4 pixel threshold. This is because human annotations still contain small localization errors and 4 pixel threshold is more reliable than 1 or 2 pixel threshold for measuring under 256x256 resolution.

\begin{table}[t]
    \centering
    \footnotesize
    \scalebox{0.96}{
    \begin{tabular}{l c c c c}
    \hline
    \textbf{Point tracking} & \textbf{All points} & \textbf{\begin{tabular}[c]{@{}l@{}}static points\\ static camera\end{tabular}} & \textbf{\begin{tabular}[c]{@{}l@{}}moving points\\ static camera\end{tabular}}  & \textbf{\begin{tabular}[c]{@{}l@{}}moving points\\ moving camera\end{tabular}}  \\ 
    \hline
    \textbf{static baseline} & 0.361 & 0.722 & 0.373 & 0.088 \\
    \textbf{TAP-Net}~\cite{doersch2022tapvid} & 0.401 & 0.496 & 0.399 & 0.328 \\
    \hline
    \end{tabular}} \\ \vspace*{6pt}
    
    \scalebox{0.97}{
    \begin{tabular}{l c c c c c c}
    \hline
    \textbf{Point tracking} & OA & $<\delta^{0}$ & $<\delta^{1}$ & $<\delta^{2}$ & $<\delta^{3}$ & $<\delta^{4}$\\
    \hline
    \textbf{static baseline} & 0.675 & 0.395 & 0.512 & 0.601 & 0.695 & 0.784  \\
    \textbf{TAP-Net}~\cite{doersch2022tapvid} & 0.849 & 0.055 & 0.214 & 0.687 & 0.927 & 0.956 \\
    \hline
    \end{tabular}} \\ \vspace*{6pt}
    
    \begin{tabular}{l c c c c c c}
    \hline
    \textbf{Point tracking} & Jac. $\delta^{0}$ & Jac. $\delta^{1}$ & Jac. $\delta^{2}$ & Jac. $\delta^{3}$ & Jac. $\delta^{4}$ \\
    \hline
    \textbf{static baseline} & 0.217 & 0.301 & 0.364 & 0.429 & 0.495 \\
    \textbf{TAP-Net}~\cite{doersch2022tapvid} & 0.025 & 0.104 & 0.442 & 0.699 & 0.734  \\
    \hline
    \end{tabular} \\ \vspace*{6pt}
    \caption{Static baseline vs TAP-Net results on the validation set. \textbf{Top}: Average Jaccard (AJ), higher is better. There are no static points in the moving camera scenario. \textbf{Middle}: Occlusion Accuracy (OA) and Position Accuracy ($<\delta^x$), higher is better. TAP-Net outperforms static baseline when measured above 4 pixel threshold. \textbf{Bottom}: Jaccard at $\delta$, higher is better. TAP-Net outperforms static baseline when measured above 4 pixel threshold.}
\label{tab:point-results-details}
\end{table}

\noindent{\textbf{Temporal action localisation:}}
We obtained baseline results for temporal action localisation using ActionFormer~\cite{zhang2022actionformer} with different pretrained features: TSP video features from~\cite{alwassel2021tsp} extracted using a Resnet(2+1)D-34 model pre-trained on ActivityNet, MMV audio features from~\cite{alayrac2020self} extracted using an S3D model pre-trained on AudioSet, and a multimodal input by concatenating the video and audio features. 
The video features have 512-dim and an effective stride of 32 (corresponding roughly to one feature per second): every other input frame is skipped and the model performs a temporal downsampling of 16. The audio features are extracted using a window length of 960ms, window stride 16000. The input audio is downsampled from 48khz to 16khz (keeping every third sample). This results in roughly 2 features per second, each of dimension 256. When using multimodal inputs, the video features are tiled over time (factor 2) to align them with the audio features.

We trained the transformer blocks and the classification and regression heads to accommodate for the number of classes included in our dataset. The resulting mean average precision is included in Table~\ref{tab:action-sounds-map}, top. The baseline struggles mostly with rare action classes and pretend actions, which are confused with their non-pretend counterpart class. Using only the audio modality leads to very poor performance, whereas using multimodal inputs does not increase the performance significantly.

\begin{figure}[h!]
    \centering
    \includegraphics[width=\textwidth]{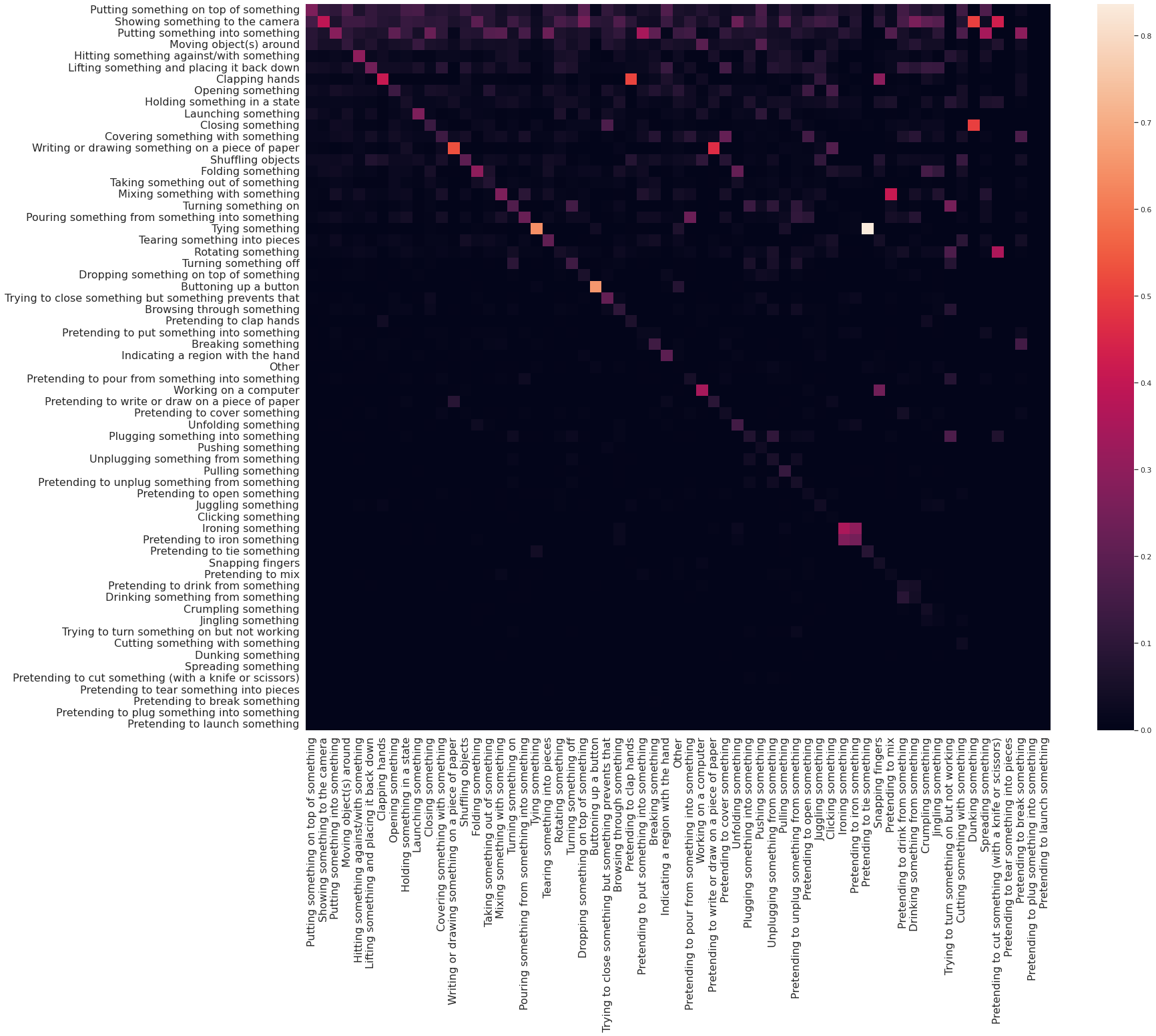} 
    \caption{Confusion matrix for ActionFormer predictions on the action localisation task. To be considered as a prediction for a certain segment, the model's confidence has to be above 0.1 and IoU threshold between the prediction and ground truth above 0.1. Ground truth actions are listed on the y-axis, sorted by their frequency; entries are normalised by rows. The less frequent actions are often confused with more frequent actions. The model also confuses pretend actions with their non-pretend versions, \eg \textit{ironing something} vs \textit{pretending to iron something} or \textit{writing or drawing something} vs \textit{pretending to write or draw}.}
    \label{fig:cm}
\end{figure}

Figure~\ref{fig:cm} shows the confusion matrix for the action localisation task, normalised by columns. It can be observed that the less frequent actions are often confused with more frequent ones and the model also confuses pretend actions with their non-pretend versions, \eg \textit{ironing something} vs \textit{pretending to iron something} or \textit{writing or drawing something} vs \textit{pretending to write or draw}.

\begin{table}[h!]
\footnotesize
\centering
\begin{tabular}{llccccccc}
\hline
&&\multicolumn{7}{c}{\textbf{Temporal Action Localisation}}\\
\textbf{Model} &\textbf{Modality} & \textbf{@0.1} & \textbf{@0.2} & \textbf{@0.3} & \textbf{@0.4} & \textbf{@0.5} &\textbf{Avg} & \textbf{\# epochs}\\
\hline
\textbf{ActionFormer} & video &17.67 &16.56 &15.13 &13.28 &11.07 &14.74 &35\\

\textbf{ActionFormer} & audio &7.25 &6.53  &5.70  &4.67  &3.64  &5.56 &55 \\
\textbf{ActionFormer} & video+audio &18.82  &17.63  &15.98  &13.99  &11.37  &15.56 &35\\
\hline
\end{tabular}\\ \vspace*{6pt}
\begin{tabular}{llccccccc}
\hline
&&\multicolumn{7}{c}{\textbf{Temporal Sound Localisation}}\\
\textbf{Model} &\textbf{Modality}& \textbf{@0.1} & \textbf{@0.2} & \textbf{@0.3} & \textbf{@0.4} & \textbf{@0.5} &\textbf{Avg} & \textbf{\# epochs}\\
\hline
\textbf{ActionFormer} & video &17.85  &15.54  &13.81  &12.11  &5.89  &13.04 &55\\
\textbf{ActionFormer} & audio &16.28 &13.58 &10.80 &8.43 &5.87 &10.99 &80\\

\textbf{ActionFormer} & video+audio &22.24 &18.99 &15.36 &11.99 &8.74 &15.46 &55 \\
\hline
\end{tabular}
    \caption{Mean average precision (mAP) for temporal action localisation (top) and sound localisation~(bottom) tasks using ActionFormer as baseline. IoU for 0.1-0.5 are averaged as in~\cite{Damen2021TheED}. \# epochs represents the number of training epochs used to obtain the best results for each experiment setup.}
    \label{tab:action-sounds-map}
\end{table}

\noindent{\textbf{Temporal sound localisation:}}
We use the same model architecture and pre-trained features as above. We trained from scratch the transformer blocks and the classification and regression heads. 
For both training and evaluation, we keep only 11 sound classes, excluding the classes corresponding to indistinguishable sounds (\eg \textit{Other:background}, \textit{Other:human}), as they hinder learning. The resulting mean average precision is included in Table~\ref{tab:action-sounds-map}, bottom. The best performance is obtained when features from both modalities are used as input. 

\begin{figure}[h]
    \centering
    \includegraphics[width=.9\linewidth]{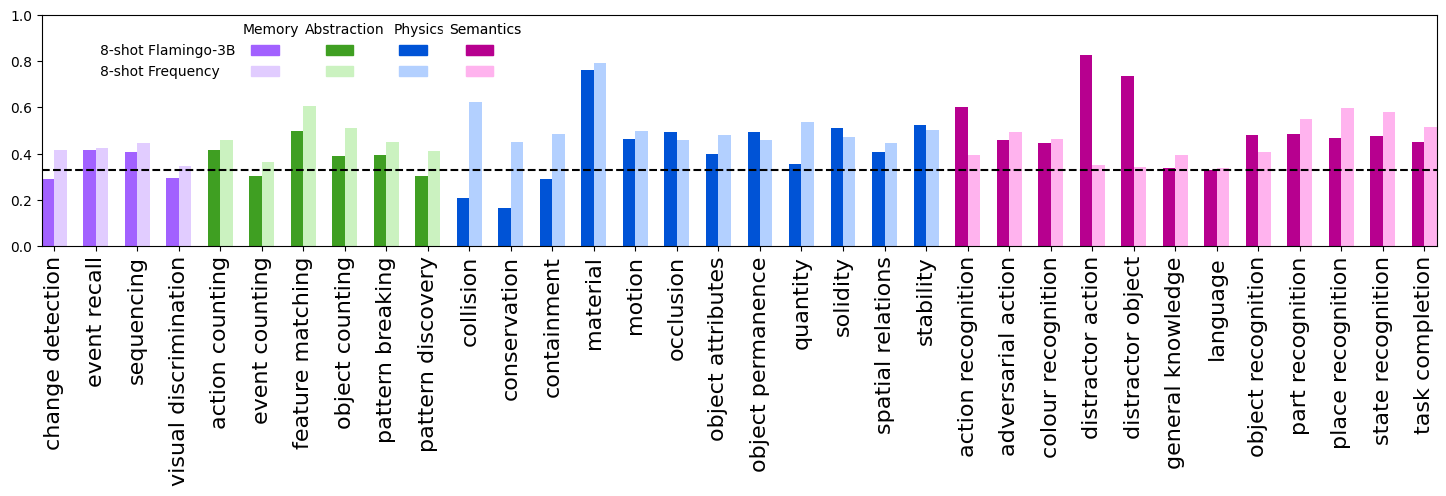}
    \caption{Performance on the validation set across skills for the 8-shot Flamingo-3B compared to 8-shot dummy frequency baseline. The black dashed line indicates the random baseline.}
    \label{fig:flamingo-skills}
\end{figure}

\noindent{\textbf{Multiple-choice videoQA:}}
For this task, we provide results for two strong recent video language models: Flamingo~\cite{alayrac2022flamingo} in zero-shot and few-shot setups, and Sevila~\cite{yu2023self} in zero-shot and fine-tuned regimes. We also include a dummy frequency-based baseline and a human baseline; see Table~\ref{tab:vqa_comparison}, Figure~\ref{fig:flamingo-human} and~\ref{fig:flamingo-skills}.

\noindent \textit{Frequency baseline}. Given that we have a fixed set of question-answer pairs defined over multiple videos, we define a simple baseline that computes how frequently each of the three options is the correct answer in the training set, and keeps the most frequent one. This baseline obtains 55.1$\%$. One can also compute this baseline on a random subset of training examples for each question type, see Table~\ref{tab:vqa_comparison}. This is a fairer dummy baseline for models using few-shot evaluation. Note that the dummy random baseline is equivalent to a 0-shot frequency baseline. The fact that the 8-shot performance of this dummy frequency baseline is above the 0-shot (random) baseline performance indicates an imbalance between the frequencies of correct answers across options in the dataset. This happens mostly because a number of questions in the dataset are binary in nature (\eg \textit{Is the camera moving or static?}), but a third option was added to comply with the 3-options-per-question setting; so in this case the possible options are: (1) \textit{moving}, (2) \textit{static or slightly shaking}, (3) \textit{I don't know}, but the third option is never or very rarely the correct one, bringing the performance of this dummy frequency baseline slightly above 50\%.

\noindent \textit{Flamingo}. We run the model with a maximum of 30 frames sampled at 1fps, spatial resolution 320. When the videos are longer than 30 seconds, we use only the middle clip. The audio modality is ignored as the original model was not trained to deal with it. The different options are scored based on likelihood. We considered zero-shot and 8-shot settings; see results in Table~\ref{tab:vqa_comparison}. In the zero-shot setting, the smaller version of the model obtains 43.6$\%$ on the test set. In the 8-shot setting, we sample 8 examples and associated ground truth responses from each question in the training set and use as prompts. The resulting accuracy is 45.8$\%$, again obtained by the smaller version of the model. The performance per skills is detailed in~\ref{fig:flamingo-skills}

\noindent \textit{SeViLA}. We run zero-shot and fine-tuned evaluation for the SeViLA model using the scripts provided by the authors~\cite{yu2023self}. SeViLA  has a Localizer and an Answerer module. It starts by sampling uniformly 32 frames from the video, out of which 4 frames are designated by the Localizer as keyframes that the Answerer uses for the final prediction. When fine-tuning, we update the weights of both the Localizer and Answerer modules using the training set from the \pt. Fine-tuning improves performance (from 46.2 zero-shot to 62.0), but this is still far from 0-shot human performance (91.4). The performance per skills is detailed in~\ref{fig:sevila-skills}.   

\begin{table}[ht]
    \centering
    \footnotesize
    \begin{tabular}{l r r r r}
    \hline
     \textbf{mc-vQA}   & \textbf{0-shot}  & \textbf{8-shot} & \textbf{All-shot} & \textbf{Fine-tuned}  \\
    \hline
    \textbf{Flamingo-3B} & 43.6 & 45.8 & - & - \\
    \textbf{Flamingo-9B} & 40.5 & 44.4 & - & - \\
    \textbf{Flamingo-80B} & 41.6 & 45.4 & - & -\\
    \textbf{SeViLA} & 46.2 & - & - & \textbf{62.0} \\
    \textbf{Frequency} & 33.3 & \textbf{51.0} & \textbf{55.1} & - \\
    \textbf{Human}& \textbf{91.4} & - & - & -  \\  
     \hline
    \end{tabular}
    \caption{mc-vQA top-1 accuracy (higher is better), for different evaluation modes and different models, including a human baseline, on the validation split. "-" refers to numbers that were not collected.}
    \label{tab:vqa_comparison}
\end{table}

\begin{figure}[h]
    \centering
    \includegraphics[width=.7\linewidth]{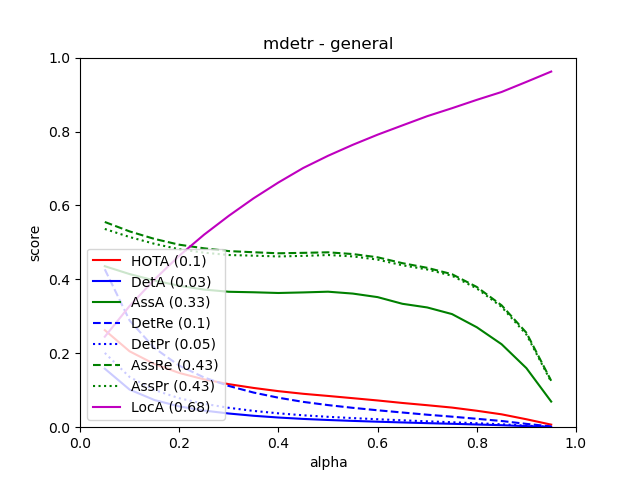}
    \caption{HOTA metrics for MDETR+Stark tracker baseline on the validation split of the \pt.}
    \label{fig:hota}
    \vspace*{-12pt}
\end{figure}

\noindent{\textbf{Grounded video question-answering:}}
In absence of a dedicated baseline in the literature for the type of grounded videoQA that we propose (input: text query, output/answer: object tracks), we obtain a simple baseline by running MDETR~\cite{kamath2021mdetr} on the middle frame of each video using the query as input, and then we use Stark tracker~\cite{Yan2021LearningST} to propagate the MDETR detections forward and backward in the video; we tried using SiamFC as tracker, but the results were worse. We measure the performance of this baseline using HOTA metrics, which integrate detection, association, and localisation scores. As expected, the performance of this baseline is poor; see Table~\ref{tab:gvqa} and Figure~\ref{fig:hota}. The failures are caused mainly by poor detection results -- since the tasks are temporal in nature, extracting \textit{seed} boxes from the middle frame is not sufficient to solve the tasks.

\begin{table}[h]
    \centering
    \footnotesize
    \begin{tabular}{l c c c c}
    \hline
    \textbf{Model} & \textbf{HOTA} & \textbf{LocA} & \textbf{DetA} & \textbf{AssA}  \\ 
    \hline
    \textbf{MDETR+Stark} & 0.1 & 0.68 & 0.03 & 0.33  \\
    \hline
    \end{tabular}
    \caption{HOTA results on the validation split for the grounded vQA task in the \pt.}
    \label{tab:gvqa}
    \vspace*{-8pt}
\end{table}

\noindent\textbf{Model size} is an essential aspect that impacts real-world applications. We report in Table~\ref{tab:params} the number of parameters of the evaluated models. For more details about the training cost of these models or inference speed, we refer to the original papers introducing these models.
\begin{table}[h]
    \centering
    \footnotesize
    \begin{tabular}{l c r}
    \hline
    \textbf{Model} & \textbf{Task} & \textbf{\# params}   \\ 
    \hline
    \textbf{SiamFC} & object tracking & 25.6M  \\
    \textbf{TAP-Net} & point tracking  &  2.8M \\
    \textbf{ActionFormer-action} & temporal action localisation  & 27.0M  \\
    \textbf{ActionFormer-sound} & temporal sound localisation & 26.5M  \\
    \textbf{Flamingo} & multiple-choice videoQA & 3B  \\
    \textbf{SeViLA} & multiple-choice videoQA & 4.1B  \\
    \textbf{MDETR+Stark} & grounded videoQA & 209M  \\
    \hline
    \end{tabular}
    \caption{Number of parameters of models evaluated on our benchmark.}
    \label{tab:params}
    \vspace*{-8pt}
\end{table}

\section{Dataset Splits Generation} \label{a:splits}
The 11.6k videos in the {\pt} are split into train, validation, and held-out test splits each with roughly $20\%/50\%/30\%$ of the videos respectively. These splits were generated by respecting two constraints:
(1)~all videos from each unique combination of (\texttt{script\_id}, \texttt{participant\_id}) are kept in the same split; more specifically, each script was filmed by a given participant possibly with multiple camera configurations, \eg from different viewpoints, or both with static and moving cameras. The above constraint ensures that all such variations of a script shot by a participant belong in the same split to avoid any leakage of video content across splits, and
(2)~various video attributes (camera motion, indoor \emph{vs.} outdoor) and annotations are divided in the same proportion across splits, \eg each split will have approximately the above specified fraction of videos with moving camera, or with point annotations. In particular, each question in the multiple-choice and grounded video QA tasks applies to a number of videos; this constraint ensures that these videos are distributed across splits in the specified proportion, such that all questions are present in all the splits.

The above was executed by setting up a linear program with a binary decision variable for each unique (\texttt{script\_id}, \texttt{participant\_id}) pair indicating which of the two splits it should be assigned to, denoted collectively ${\bf{x}}{\in} \{0,1\}^n$ with $n$ being the number of such unique pairs. Note for splitting into three splits, the problem is solved twice sequentially.
A feature count matrix $A{\in}\mathbb{R}^{n\times d}$ was constructed, with $A_{ji}$ being the number of videos shot by the $j^{th}$ (\texttt{script\_id}, \texttt{participant\_id}) having the $i^{th}$ video-attribute ($d$ being the total number of video attributes). An ``attribute'' indicating the total number of videos with a given (\texttt{script\_id}, \texttt{participant\_id}) was also included to enforce the number of videos in each split.  The following linear program was solved using the CVXPY interface to the MOSEK mixed-integer solver.
\begin{align*}
    \hfill \min_{\bf{x}} \left[\left(\max_{i} \left(1 - t_i \right)^2\right) + \frac{1}{d}\sum_{i=1}^{d} \left(1 - t_i\right)^2 \right]\hfill\\
\text{s.t.,}\hspace{5mm}  t_i = \frac{A_i^T\bf{x}}{\lceil f_1 A_i^T\mathds{1}\rceil}, \forall{i} \in \{1,\hdots,d\} \\
(1 - \lambda) \leq t_i \leq (1+\lambda), \forall{i} \in \{1,\hdots,d\}\\
\text{and,}\hspace{5mm}    {\bf{x}}_j \in \{0,1\}, \forall{j} \in \{1,\hdots,n\}
\end{align*}
with $A_i$ being the $i^{th}$ column of $A$, $f_1 \in [0, 1]$ being the target fraction for the split corresponding to label ${\bf{x}}_j = 1$ (\eg $f_1 = 0.5$ for a 50\% test split), and $\lambda=0.25$ is the maximum allowed fractional deviation from the target value. There were $n=7288$ unique (\texttt{script\_id}, \texttt{participant\_id}) pairs, and $d=249$ video attributes.

\section{Annotation collection}

\noindent{\textbf{Raters instructions:}} We include below the high-level instructions provided to raters when collecting the different types of annotations.

\noindent{\textit{Object tracks:}} Annotate, using spatio-temporal bounding box tracks, all the objects that the person interacts with. Annotate also the objects in the immediate vicinity of the objects that the person interacts with, as they act as distractor objects. In addition, annotate 3-5 objects in the background. Once you mark an object in a frame, the tracker running in the background will generate proposals for that object throughout the video. Please check every 30th frame and amend the proposals if they are not correct. For each object track, provide a representative name including object class, object attributes, \eg red mug. As much as possible, include annotations for liquids as well. When objects get torn (\eg a salad leave, a piece of paper) or are broken (\eg eggs), the object tracks and names should reflect the change in object state: \eg a single object track named "egg" would be split into multiple object tracks named "egg-shells" and "egg-content" once the person breaks the egg. For objects that go out of the field of view and reappear later on, make sure to assign the same object track.

\noindent{\textit{Point tracks:}} Given a video with an inpainted box track, select and track at least 3 points inside the object, belonging to different parts of the object. Mark the start and end points of the track, then wait for the optical flow estimator running in the background to provide predictions for the intermediate frames. Check and correct any errors you notice on all the intermediate frames. Assign names to each point corresponding to the semantic part to which the point belongs to. When a point becomes occluded (because of object rotation or object going out of the field of view), mark the point as \textit{occluded}.  

\noindent{\textit{Action segments:}} Given the list of templated action labels (\eg putting \textit{something} into \textit{something}), mark the start and end points of each action segment. As action boundaries, please use the moments of contact with the objects involved in the action, \eg when a person stirs a tea with a spoon, mark as start moment the moment when the person picks up the spoon, and as end moment the moment when the person stops stirring or puts down the spoon. In addition to indicating the action boundaries, select from the existing object tracks the tracks involved in the action segment, in the order in which they appear in the template, \eg when a person pours water from a kettle into a cup, mark the segment as \textit{pouring something from something into something}, and indicate \textit{water}, \textit{kettle}, \textit{cup} as relevant objects, in this order. If the person repeats the same action multiple times (\eg clapping hands), mark separate segments as much as possible.   

\noindent{\textit{Sound segments:}} Given the list of sound labels (\eg clapping hands, hitting something), mark the start and end points of each sound segment. In addition, select from the existing object tracks the tracks involved in producing the sound, \eg when the person drops a cable on a table, mark the sound as \textit{hitting something}, and select \textit{cable}, \textit{table} as objects involved in producing the sound.  

\noindent{\textit{Multiple-choice videoQA:}} 
\textit{Phase 1} (open-ended questions): Answer the following questions about this video using short sentences written in English. \textit{Phase 2} (multiple-choice questions): Answer the following questions about this video by selecting the correct answers from the given ones. Only one option is correct for each question. 

\noindent{\textit{Grounded videoQA:}} The answers to questions were generated automatically from the object tracks annotations above, using simple heuristics. These annotations were then checked by human raters with the instruction: Given the question or query below and the video with one or more inpainted object tracks, indicate (yes/no) if the inpainted objects correctly answer the question.   

\noindent{\textbf{Data collection pipelines:}} The different types of annotations were collected using two different approaches:
\begin{enumerate}
    \item \textit{sequential pipeline} for the object and point tracks, action and sound segments: (i) a rater annotates a video for a given task, (ii) a second rater checks the annotation, makes any necessary corrections, then marks the annotation as complete; (iii) a third rater checks if the annotation is indeed complete or it needs additional changes, in which case they will send the video back to step (ii) to be reviewed by a different rater.
    For difficult tasks like point tracking or object tracking with hard occlusions, we did multiple annotation cleaning rounds, each time with specific cleaning guidelines. For example, for the videos in cups-games category mentioned above, in one cleaning round, the raters were asked to pay attention to the hidden object, or for videos where the person shows objects to the camera sometimes repeating the same object, we asked raters to pay attention to assign the same object ID when the object reappears. Having videos grouped by script type helped in designing specific cleaning guidelines to ensure good annotation quality.  
    \item \textit{parallel pipeline} for multiple-choice and grounded videoQA: multiple raters answer in parallel the same question for the same video and the option chosen by the majority of raters is kept as final answer. Note that for multiple choice QA, during annotation collection, the raters were presented with more than 3 options in some cases. For the final dataset, as the goal was to have the same number of options for all the questions, we chose to keep 3 options to accommodate binary questions as well (where the options used are: \textit{Yes}, \textit{No}, \textit{I don't know}). For questions with more than 5 options, the negative options were sampled based on their frequency as correct options for videos in the same script type. Finally, for some generic questions, \eg \textit{Which statement describes the scene better?}, the answers were collected initially in open-language format, and then negatives were sampled using the answers from other videos in the same script type, with additional checks from the research team to avoid ambiguous distractors. 
\end{enumerate}

\begin{figure}[h]
    \centering
    \includegraphics[width=\linewidth]{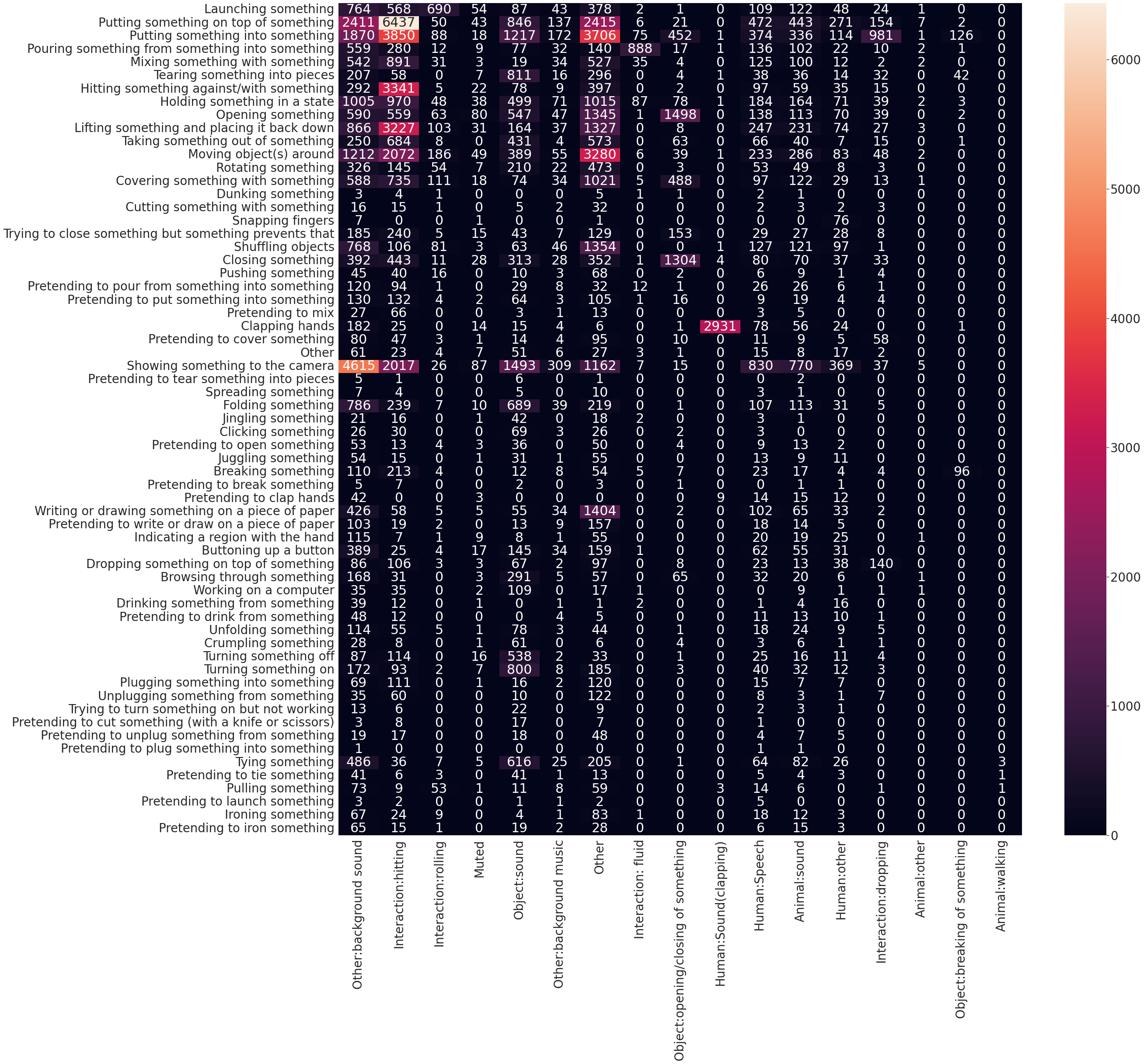}
    \caption{Correlation between action and sound temporal annotations in the \pt.}
    \label{fig:action-sound}
\end{figure}

As a sanity check, for the action and sound annotations, we checked for overlapping objects involved in both action and sounds (see Figure~\ref{fig:action-sound}). We observed strong correlations across pairs of action-sound, indicating consistent annotations across modalities, \eg the \textit{Pouring something into something} action shares the same objects with the \textit{Interaction: Fluid} sound, the  \textit{Clapping hands} action co-occurs with the  \textit{Human (clapping)} sound, the  \textit{Lifting something and putting it back down} action co-occurs with the \textit{Object: Hitting} sound, \textit{Moving something around} actions co-occurs with \textit{Object: Rolling} sound, and so on.

\section{Compensation} 
Besides the research team creating the benchmark, we relied on three groups of contributors: participants filming the videos, raters annotating the videos, and human participants involved in the human baseline collected for the multiple-choice vQA task. The full details of our study design, including compensation rates, were reviewed by DeepMind’s independent ethical review committee. All participants provided informed consent prior to completing tasks and were reimbursed for their time. The policy ensures that workers/participants are paid at least the living wage for their location.

\section{Diversity of participants involved in filming}
We consider that good representation of the world's population in terms of different demographics is an essential aspect in benchmarking multimodal models, to ensure a safe and fair deployment of such models world-wide. When building our benchmark, we considered three diversity aspects for the participants involved in filming: gender, ethnicity, and country of residence. These factors offer visual diversity in the dataset in terms of appearance of people and scenes. We acknowledge that this is not a complete coverage of diversity factors, and other aspects such as age, disability, household income, or level of education are important to control, and we hope to be able to include such factors in future iterations of the benchmark.

\begin{figure}[h]
    \centering
    \includegraphics[width=\textwidth]{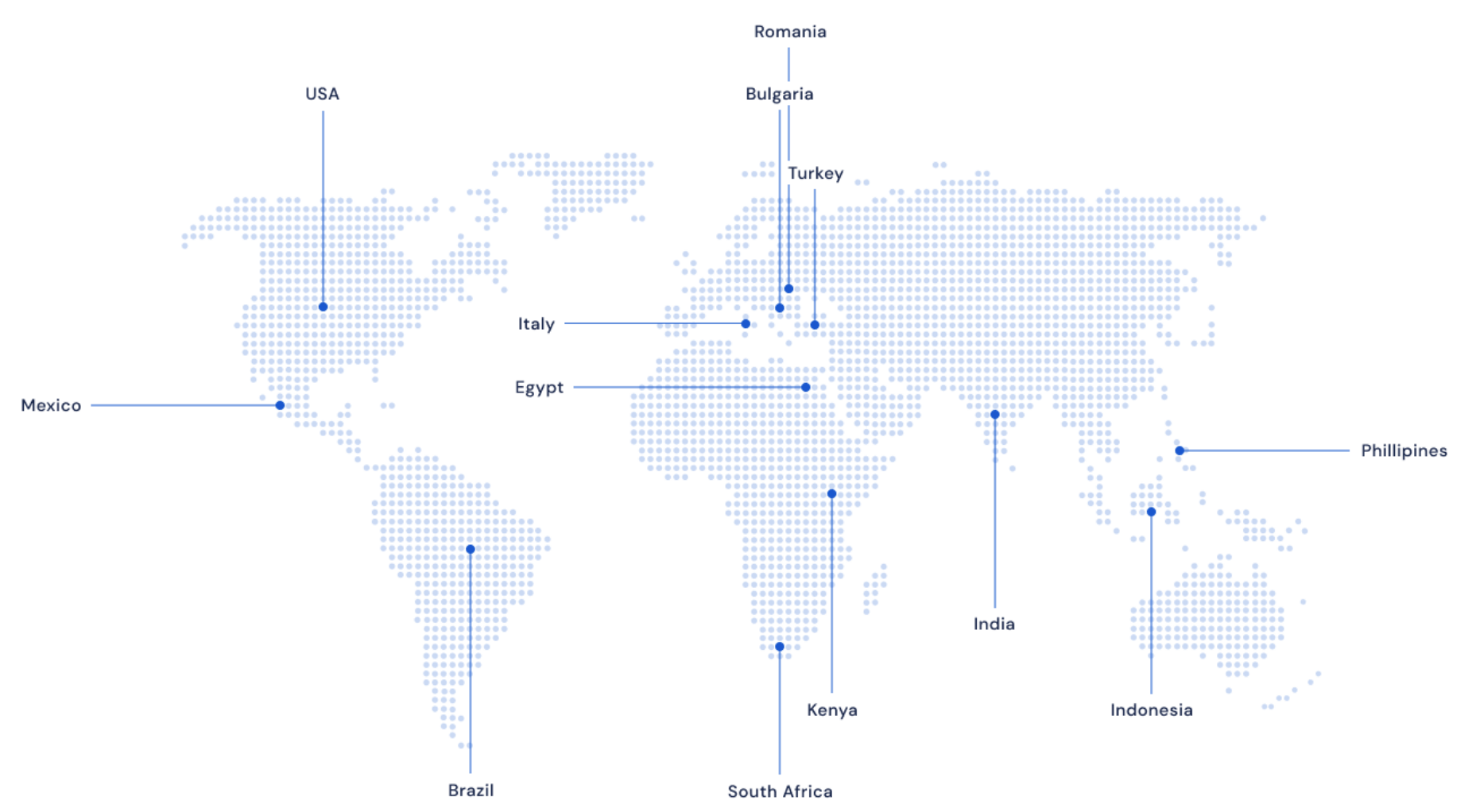}
    \caption{Geolocation of participants involved in filming.}
    \label{fig:country}
    \vspace*{-12pt}
\end{figure}

We include in Table~\ref{tab:diversity} and Figure~\ref{fig:country} the self-reported demographics along these axes. Note that all the demographic information was self-reported by the participants themselves. It can be observed that there is a good balance across gender and good spread across ethnicity (providing diversity in terms of skin-tone). Filming the videos in more than 13 countries on different continents provides good scene and objects variation.

\begin{table}[h]
    \centering
    \footnotesize
    \small{
    \begin{tabular}{c c}
    \hline
         \textbf{Gender} & $\%$ \\
         \hline
         Male & 46.40 \\
         Female, Other & 53.60  \\ 
         \hline
         \hline
         \textbf{Ethnicity} & $\%$ \\
         \hline
         White or Caucasian & 28.97 \\
         South and East Asian & 25.49  \\ 
         Black or African American & 21.68 \\
         Latino or Hispanic & 9.25 \\
         Mixed & 3.94\\
         Middle Eastern & 3.37 \\
         Other & 7.30 \\
         \hline
    \end{tabular}
    \vspace{1mm}
    \begin{tabular}{c c}
    \hline
         \textbf{Country} & $\%$ \\
         \hline
         Philippines & 31.38 \\
         Brazil & 11.27  \\ 
         Kenya & 10.02  \\
         Indonesia & 8.75 \\
         Italy & 8.03 \\
         Romania & 7.57 \\
         South Africa & 5.25 \\
         Turkey & 4.12 \\
         India & 3.72 \\
         Mexico & 1.45 \\
         Bulgaria & 1.37 \\
         United States & 0.70 \\
         Egypt & 0.48 \\
         Other & 5.87 \\
         \hline
    \end{tabular}
    }
    \caption{Self-reported demographics (Gender, Ethnicity, Country) of participants involved in filming.}
    \label{tab:diversity}
    \vspace*{-8pt}
\end{table}

\section{Limitations and potential negative societal impact}
\label{sec:limitations}

\noindent\textbf{Limitations:}
We designed video scripts and questions to have a broad coverage of perception skills and types of reasoning, across different modalities (video, audio, text), probed through high-level and low-level computational tasks. Given the broad coverage, it was challenging to have a perfect balance across all dimensions. As future work, we aim to add more tasks that require counterfactual reasoning or memory skills, and more annotations for grounded vQA and point tracking. In addition, the balance across options in the multiple-choice videoQA is not perfect, as indicated by the fact that the frequency dummy baseline obtains better performance compared to the pure random baseline (55.1\% vs 33.3\%). When analysing model performance, we need to take such biases into account. 

Our benchmark aims to comprehensively evaluate multimodal models' performance across different perception skills. However, some modalities are missing, \eg touch, or some aspects are not covered by the available annotations, \eg force, deformations, detailed 3D geometry. We will work to improve the coverage in future iterations, and we also welcome contributions from the community to add more tasks, modalities, even new languages to the \pt. Note, however, that our benchmark focuses on temporal tasks defined over videos. Many current multimodal models can only handle image and text as modalities, hence it might not be straightforward to evaluate them on our benchmark. We do not consider this to be a limitation of our benchmark, but a limitation of those models, as we believe that general multimodal perception models need to be able to perceive and reason over spatial and temporal dimensions, across modalities. 

As mentioned in the main paper, most of the videos were filmed on a table-top, using common household objects, following the scripts designed by our research team. This could be perceived as a setup with limited diversity when compared to “in the wild” videos available in repositories like Youtube. However, we argue that for evaluating a general perception model, it is important to isolate the skills and types of reasoning we care about, while building in invariance to lighting, camera angle, types of objects, the person’s skin tone, etc – these are obtained by filming the same script (with multiple variations) with 20+ different participants per script variation, who choose on their own where to place the camera, what exact type of object to use for a certain action, in what order to perform some actions, etc. Curating videos in the wild to obtain the same coverage of skills and types of reasoning would be hard, even impossible, since some types of data simply don’t exist online in sufficient numbers (\eg correct vs. incorrect execution of actions).

While filming, we instructed participants to use 2 different viewpoints to obtain more diverse camera angles. However, this information is not explicitly used at the moment in our tasks (we only used this information when deciding the splits, to make sure these paired videos fall in the same split). While some participants filmed simultaneously with two cameras, others recorded two runs one after the other. This approach was previously used in crowd-sourced datasets (e.g. Charades-Ego dataset). Through manual inspection, we note that the variations are minor as the sequences were recorded one after the other. Such paired videos could be useful to design new tasks.

Related to missing capabilities, our benchmark cannot accommodate \textit{active} perception evaluation. Enabling interaction would limit us to simulated environments. To still address the agency aspect to some extent, we included videos where the model is required to recognise correct and incorrect execution of certain actions (e.g. tying shoe laces, buttoning up a shirt, covering a container with a cover, pouring water in a glass) or to assess the consequences of actions (e.g. what would happen if we remove a certain object from the table) – these are possible because our scripts include multiple variations with correct/incorrect actions, or controlled variations of object configurations, which would be impossible to curate from public repositories like Youtube.

As mentioned in the main paper, our benchmark is medium scale, comparable to Charades, but an order of magnitude smaller compared to \eg Ego4D. We would like to emphasise that this is not a limitation of the benchmark since our focus is on evaluation; large-scale datasets are needed for training. We provide a medium-scale dataset with a small set for fine-tuning / prompting, and the rest for evaluation, as in this way we can probe the generalisation power of multimodal (pre-trained) models.  

\noindent\textbf{Societal impact:} 
Benchmarks such as ours guide research and, indirectly, lead to improvement in models' capabilities. These models can be used for many different applications that have positive societal impact, \eg video-language models assisting the visually-impaired, or surveillance systems for the elderly or children. However, similar systems can be used to cause harm (\eg intrusive surveillance systems). We hereby state our strong stance against the use of our benchmark for evaluating and improving such systems.

\end{document}